\documentclass[11pt]{article}

\usepackage[final]{acl}

\usepackage{times}
\usepackage{latexsym}

\usepackage[T1]{fontenc}

\usepackage[utf8]{inputenc}

\usepackage{microtype}

\usepackage{inconsolata}

\usepackage{graphicx}
\usepackage{subcaption}
\usepackage{epigraph}

\usepackage{tcolorbox}
\usepackage[utf8]{inputenc}
\usepackage{graphicx}
\usepackage{booktabs}
\usepackage[table]{xcolor}
\usepackage{soul}
\usepackage{multirow}
\usepackage{amssymb}
\usepackage{amsmath}
\usepackage{tabularx}
\usepackage{enumitem}
\usepackage{cleveref}
\usepackage{twemojis}
\usepackage{xspace}
\usepackage{makecell}
\usepackage{array}

\newcommand{\gray}[1]{\textcolor{gray}{#1}}
\definecolor{gray94}{gray}{.94}
\definecolor{gray90}{gray}{.90}
\definecolor{lightblue}{RGB}{221,235,247}

\newcommand{\grow}[1]{\rowcolor{gray94}{#1}} 
\newcommand{\brow}[1]{\rowcolor{lightblue}{#1}} 

\newcommand{\dy}[1]{#1}

\newcommand{\ourmethod}{\textsc{IRiS}\xspace}

\crefformat{section}{#2\S~#1#3}
\Crefformat{section}{#2\S~#1#3}

\title{Beyond Static Personas: Situational Personality Steering for Large Language Models}

\author{
  Zesheng Wei$^1$\thanks{Equal contribution.}\textsuperscript{,}\thanks{Work was done during a visit at SMU.} \quad
  Mengxiang Li$^1$\footnotemark[1] \quad
  Zilei Wang$^1$\thanks{Corresponding author.} \quad
  Yang Deng$^2$ \\
  $^1$University of Science and Technology of China \\
  $^2$Singapore Management University \\
  \texttt{\{zswei, mxli02\}@mail.ustc.edu.cn}, \texttt{zlwang@ustc.edu.cn} \\
  \texttt{ydeng@smu.edu.sg}
}

\begin{document}

\maketitle

\begin{abstract}
Personalized Large Language Models (LLMs) facilitate more natural, human-like interactions in human-centric applications. However, existing personalization methods are constrained by limited controllability and high resource demands. Furthermore, their reliance on static personality modeling restricts adaptability across varying situations. To address these limitations, we first demonstrate the existence of situation-dependency and consistent situation-behavior patterns within LLM personalities through a multi-perspective analysis of persona neurons. Building on these insights, we propose \ourmethod, a training-free, neuron-based Identify\mbox{-}Retrieve\mbox{-}Steer framework for advanced situational personality steering. Our approach comprises situational persona neuron identification, situation-aware neuron retrieval, and similarity-weighted steering. We empirically validate our framework on \textsc{PersonalityBench} and our newly introduced SPBench, a comprehensive situational personality benchmark. Experimental results show that our method surpasses best-performing baselines, demonstrating \ourmethod's generalization and robustness to complex, unseen situations and different models architecture.

\end{abstract}
\section{Introduction}
\label{sec:introduction}

\epigraph{\textit{``Behavior is a function of the person and their environment.'' \citep{lewin2013principles}}}{--- Kurt Lewin}

\noindent The advancement of Large Language Models (LLMs) has catalyzed a wide range of human-centric applications such as role-playing \citep{roleplay-survey,wang2025cosercoordinatingllmbasedpersona, wang-etal-2025-characterbox}, personalized assistance \citep{sigir24-pca,mok-etal-2025-exploring}, user simulation \citep{emnlp24-trip,emnlp25-persona-esc}, and social simulation \citep{ICLR2024_578e65cd, zhou2024sotopiainteractiveevaluationsocial,multilingual-socialsim}. 
\dy{These applications require models to express coherent personalities while adapting their behavior across diverse interaction situations, making personality modeling a central challenge for LLM personalization. However, most existing approaches implicitly assume that behavior is determined solely by stable personality traits. This assumption conflicts with a core principle in psychology: \textit{behavior is a function of both the person and the environment} \cite{lewin2013principles}. When situational factors are ignored, personalized agents may exhibit superficial consistency yet fail to respond appropriately across varying situations.}

\begin{figure}[t]
\setlength{\abovecaptionskip}{5pt}   
\setlength{\belowcaptionskip}{0pt}
    \centering
    \includegraphics[width=\linewidth, trim=8pt 8pt 889pt 767pt, clip]{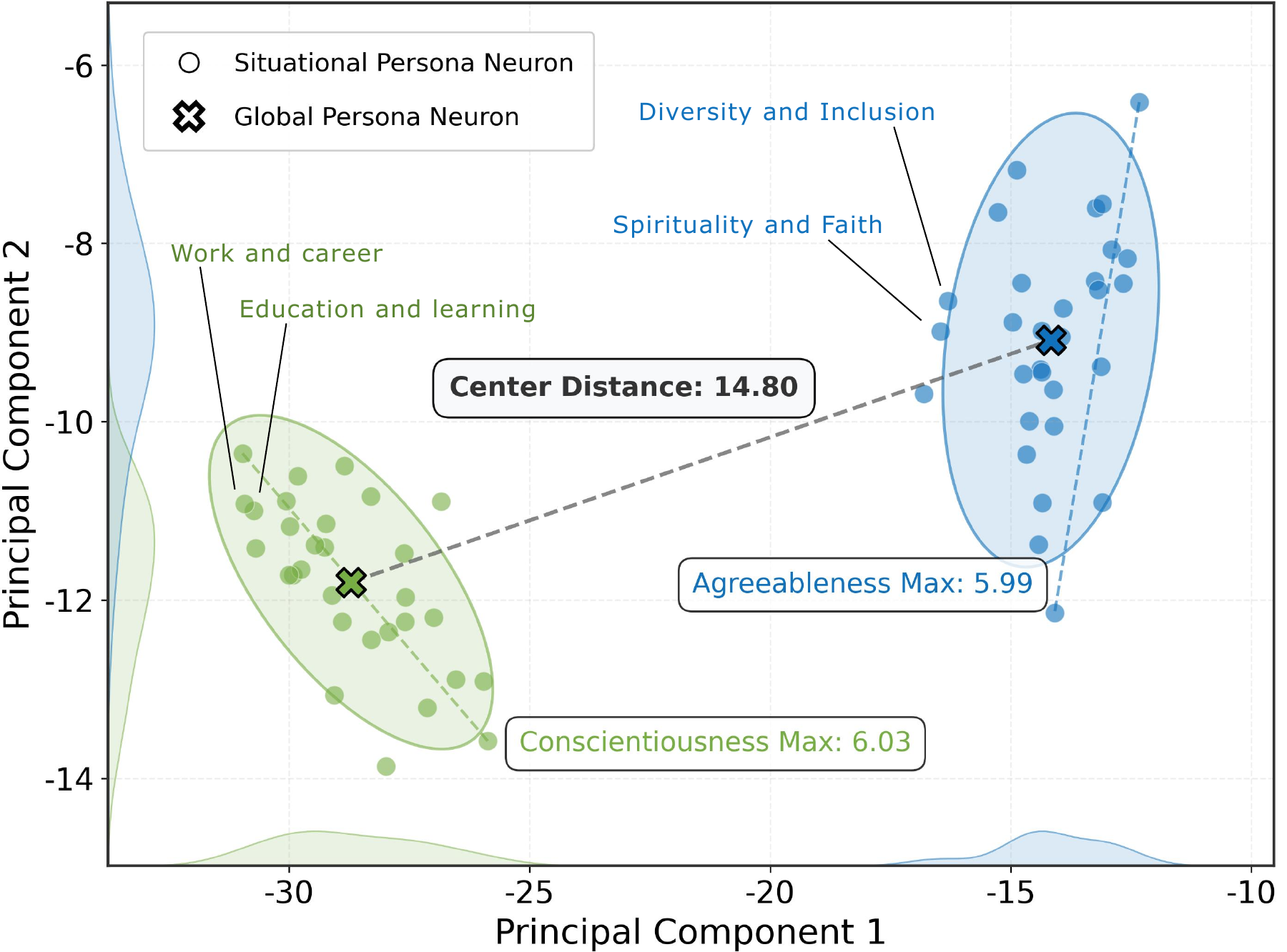} 
    \caption{PCA of situation-dependent and global persona neurons, with proximate topic names annotated. Comparable distances between distinct personality domains and topics within same personality domain highlight the crucial impact of situations.
    \label{fig:PCA_qwen_appendix}
}
\vspace{-3mm}
\end{figure}

Existing approaches to endowing LLMs with personality are primarily categorized into training-based and training-free methods. Training-based methods rely on large-scale, high-quality datasets to align models with specific personalized preferences \citep{li-etal-2025-big5}, utilizing techniques such as Supervised Fine-Tuning (SFT) \citep{wang2025opencharactertrainingcustomizableroleplaying, tan-etal-2024-democratizing, li2024personalizedlanguagemodelingpersonalized} or Direct Preference Optimization (DPO) \citep{li2024evaluatingpsychologicalsafetylarge}. 
\dy{While effective, these methods are computationally expensive \citep{szep2025finetuninglargelanguagemodels} and difficult to adapt post hoc \citep{tseng-etal-2024-two}.}
\dy{Training-free methods, including prompt-based personalization \citep{10.5555/3666122.3666588, li2023largelanguagemodelsunderstand} and direct internal steering \citep{deng2024neuronbasedpersonalitytraitinduction, chen2025personavectorsmonitoringcontrolling}, offer greater flexibility but suffer from instability, limited controllability, and weak theoretical grounding.}
Although prompt-based methods can incorporate conversational history as situational context, they treat it merely as a "black-box" input. Relying on implicit attention mechanisms for this modulation lacks transparency, often causing unstable personality expression. Crucially, both training and training-free paradigms lack a mechanistic framework to explicitly model the underlying persona-situation interactions.


\dy{In contrast to the static assumptions underlying existing LLM personalization methods, personality psychology provides a well-established \textbf{persona\mbox{-}situation interactional} account of behavior \cite{lewin2013principles}. While early trait theories assume cross-situational consistency \citep{newcomb1929consistency, allport1937personality}, subsequent work demonstrates that stable traits alone are insufficient to explain behavior across diverse situations \citep{mischel1968personality, mischel1982beyond}. 
In particular, the Cognitive-Affective Personality System (CAPS) theory \citep{mischel1995cognitive} argues that \textit{human behavior varies across diverse situations}. Empirical studies further demonstrate that \textit{different situations selectively activate cognitive and affective units} \citep{mischel2002situation}, and that \textit{personality coherence is expressed through consistent situation\mbox{-}behavior patterns} \citep{ayduk2008applying}. To date, these persona\mbox{-}situation interactional mechanisms have not been systematically investigated or integrated into LLM-based personalization.}


\dy{Motivated by the above psychology studies, we first empirically examine whether LLMs exhibit human-like situational dependency and consistent situation\mbox{-}behavior personality patterns, through multi-perspective analysis of internal neurons. As illustrated in Figure~\ref{fig:PCA_qwen_appendix}, our preliminary results show that the maximum PCA distance between situational topics within a single domain is comparable to the distance observed between distinct personality domains, suggesting that situational variation can induce significant personality-level representational shifts. Building on this observation, we propose a training-free, neuron-based Identify\mbox{-}Retrieve\mbox{-}Steer framework for situational personality steering, named \ourmethod.
The framework first identifies situational persona neurons from a set of historical situations, which serve as instructive priors. Given a novel situational situation, \ourmethod estimates its similarity to these historical situations, retrieves the corresponding persona neurons, and applies coefficient-weighted steering to enable precise and situation-aware personalization.}

Comprehensive evaluations across two personality benchmarks validate the SOTA performance of \ourmethod, demonstrating effective generalization to unseen situations and robustness within complex situations. Furthermore, extensive experiments verify the adaptability of our approach across diverse model architectures. To summarize, our contributions are as follows:
\begin{itemize}[leftmargin=*,nosep]
\item We empirically validate human-like situational dependency and consistent situation-behavior personality patterns within LLMs, pioneering the integration of these psychological mechanisms to guide precise personality steering.
\item We propose the \ourmethod framework, a psychology-grounded approach that leverages instructive priors for situational-aware retrieval and steering.
\item We conduct extensive experiments and in-depth analyses to verify our framework's effectiveness, offering intuitive insights into situational personality 
for future research and applications.
\end{itemize}

\begin{figure*}[t]
    \setlength{\abovecaptionskip}{3pt}   
    \setlength{\belowcaptionskip}{0pt}
    \centering
    
    \begin{subfigure}[b]{0.402\textwidth}
        \centering
        \includegraphics[width=\linewidth, trim=2pt 796pt 760pt 2pt, clip]{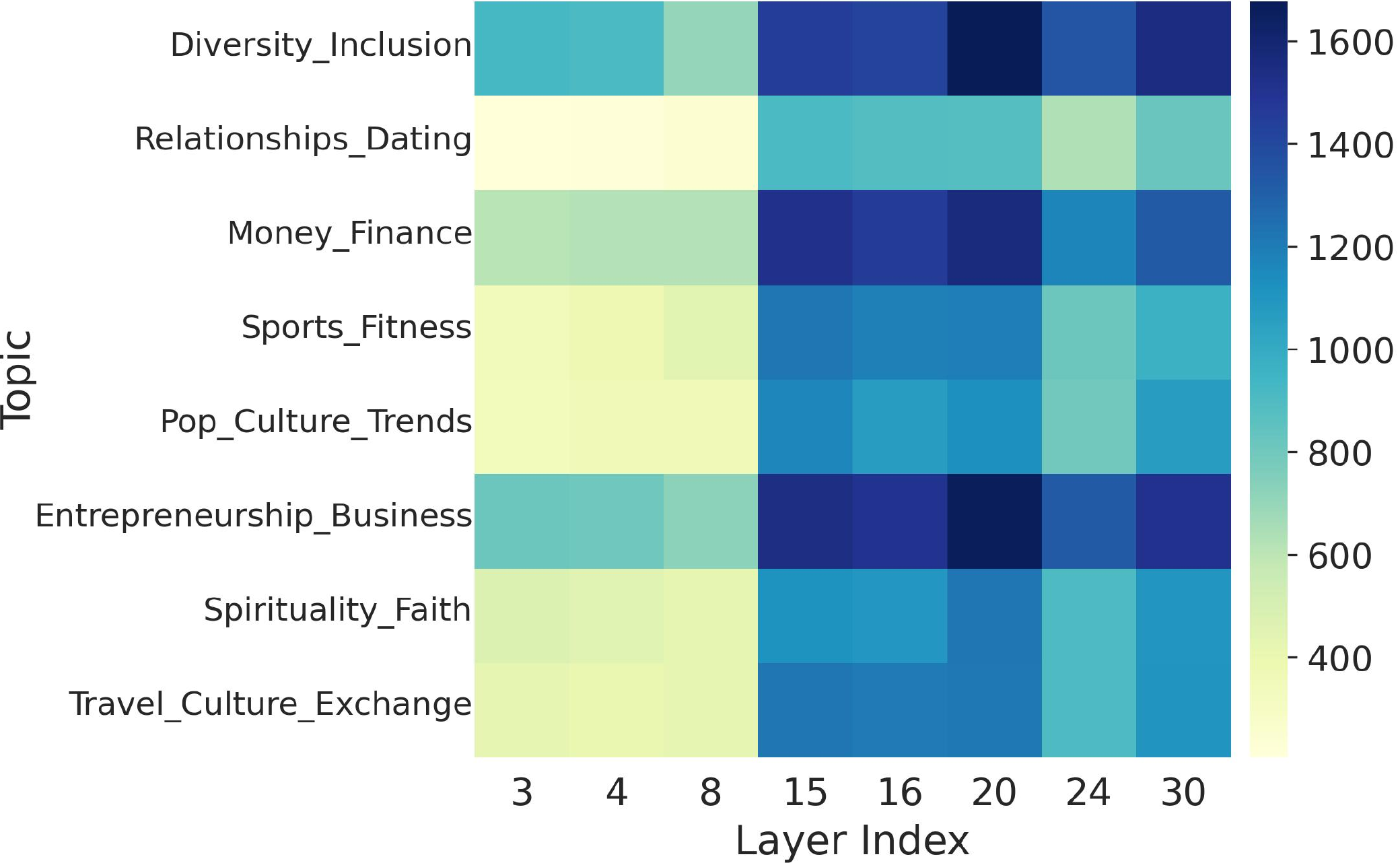}
        \caption{}
        \label{fig:heatmap}
    \end{subfigure}
    \hfill 
    \begin{subfigure}[b]{0.29\textwidth}
        \centering
        \includegraphics[width=\linewidth, trim=1pt 796pt 1040pt 2pt, clip]{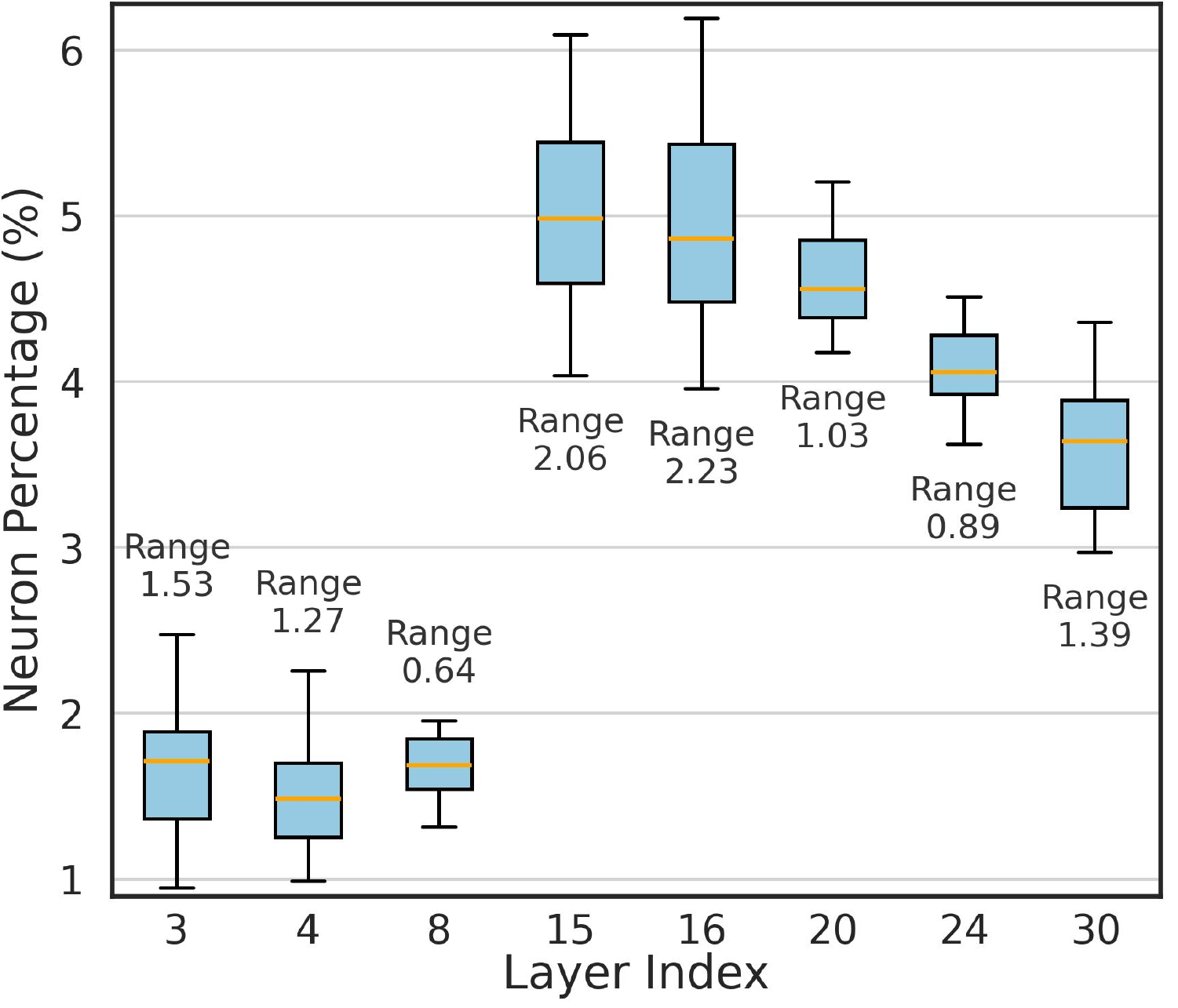}
        \caption{} 
        \label{fig:box} 
    \end{subfigure}
    \hfill 
    \begin{subfigure}[b]{0.295\textwidth} 
        \centering
        \includegraphics[width=\linewidth, trim=3pt 805pt 1047pt 11pt, clip]{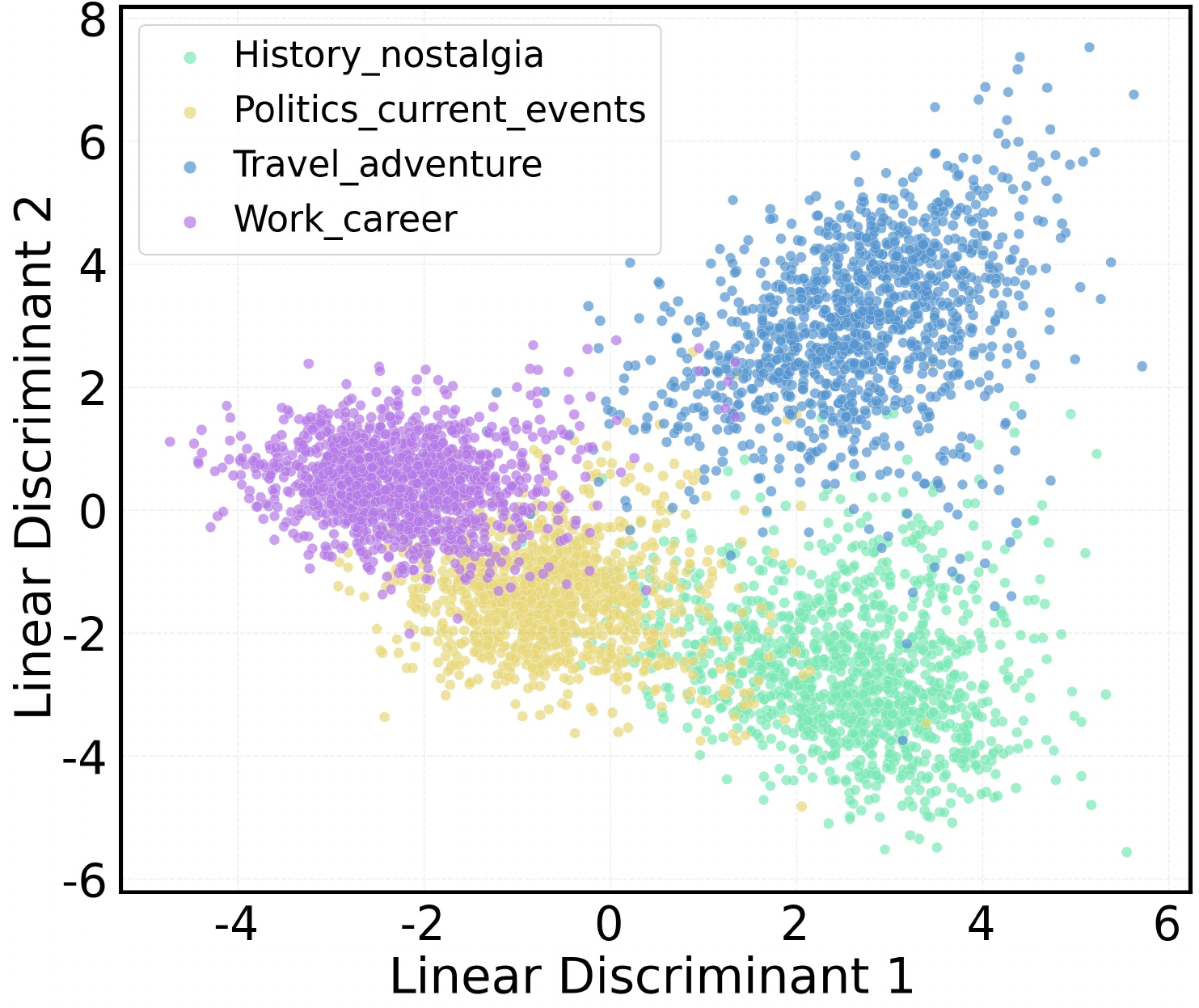}
        \caption{}
        \label{fig:PCA_LDA}
    \end{subfigure}
    \caption{Empirical study results: (a) Layer-wise counts of persona neurons across situational topics (Topic labels abbreviated by omitting "and") (b) Variability in neuron proportions across early, middle, and late layers for different topics. (c) Validation of the situation-behavior consistency patterns in LLM personality.}
    \label{fig:validation_abc}
    \vspace{-3mm}
\end{figure*}

\section{\dy{Preliminary Analysis}}
\dy{The preliminary analysis aims to investigate the situational impact on LLMs' personality manifestation, and to validate the theory of situation-behavior consistency within LLMs.}
\subsection{Backgrounds}
\paragraph{Personality Model} 
In this work, we adopt the widely validated Big-Five model \citep{Tupes1992RecurrentPF} as the foundational personality framework, which comprises five domains: \textit{\underline{o}penness} (\textbf{O}), \textit{\underline{c}onscientiousness} (\textbf{C}), \textit{\underline{e}xtroversion} (\textbf{E}), \textit{\underline{a}greeableness} (\textbf{A}), and \textit{\underline{n}euroticism} (\textbf{N}).  Each domain encompasses opposing aspects (e.g., extraverted versus introverted within the \textbf{E} domain).

\paragraph{Neurons in LLMs}
Modern LLMs adopt the auto-regressive transformer \citep{10.5555/3295222.3295349} architecture, which consists of L stacked transformer blocks. Previous work has shown that knowledge, such as personality, is stored in specific neurons within the feed-forward networks (FFNs) of each block \citep{dai2022knowledgeneuronspretrainedtransformers}. Specifically, in layer \(l\), given input \(X^l\) for a token, we have:
\begin{equation}
\small
\text{FFN}(X^l) = \text{act}(X^l W^l_1)W^l_2
\end{equation}
where $X^l \in \mathbb{R}^d$, $W^l_1 \in \mathbb{R}^{d \times d_h}$, \(W^l_2 \in \mathbb{R}^{d_h \times d}\), and \(act\) represents the activation function (e.g., ReLU \citep{agarap2019deeplearningusingrectified}). More recent advanced LLMs \citep{grattafiori2024llama3herdmodels, gemmateam2024gemmaopenmodelsbased, yang2025qwen3technicalreport} have replaced the ReLU non-linearity with the GLU \citep{shazeer2020gluvariantsimprovetransformer} activation function to achieve better performance:
\begin{equation}\small
\text{FFN'}(X^l) = (\text{act}(X^l W^l_1) \odot (X^l W^l_3))W^l_2
\end{equation}
where \(\odot\) denotes element-wise multiplication, and \(W^l_3 \in \mathbb{R}^{d \times d_h}\) is the gating weight matrix.

In layer \(l\), the \(i^{th}\) neuron can be conceptualized as applying a linear transformation to column \(i\) of \(W^l_1\) followed by the non-linear activation. The activation value of a neuron is positively correlated with the expression of corresponding facts. In this context, a neuron is considered \textbf{activated} if its activation value exceeds zero \citep{10.5555/3104322.3104425}.

\paragraph{Neuron Identification Dataset \& 
\dy{Situational Topic} Category} 
\label{sec:Neuron_Identification_Dataset}
To identify persona neurons within LLMs, we employ the dataset from \textsc{PersonalityBench}  \citep{deng2024neuronbasedpersonalitytraitinduction}, denoted as $\mathcal{Q}$, which comprises a diverse set of descriptive personalization prompts and situational questions designed to elicit personality-driven responses. To investigate the influence of varying situations on LLMs' personality, these questions are categorized  into \(M\) = 30 distinct topics, following the taxonomy of UltraChat \citep{ding-etal-2023-enhancing}. Further details are provided in Appendix~\ref{appendix:topic_set}.

\subsection{Empirical Study}
\label{sec:emp_study}
\dy{The targets of our analysis} are \textit{\textbf{situational persona neurons}} \dy{in LLMs}, which are identified by measuring activation differences of neurons within LLMs under contrastive personality prompts within 
\dy{situational} topics, with detailed explanations in \cref{sec:identify_neurons}.



\subsubsection{Situational Impact on LLMs' Personality}
\label{sec:empiri_situation_impact}
We analyze the activation states and layer-wise distribution of persona neurons to investigate situation impact on LLMs' personality. 
Llama-3-8B-Instruct \citep{grattafiori2024llama3herdmodels} and Qwen3-8B \citep{yang2025qwen3technicalreport} \dy{are adopted} as target LLMs \dy{for validation}. Comprehensive results are provided in Appendix \ref{appendix:complete_empirical_study}.

\paragraph{Activation-Level Perspective}  
We collect the activation probabilities of all situational persona neurons across diverse topics. For comparison, we also derive "global persona neurons" by disregarding topic distinctions, establishing a baseline for cross-situational consistency. We then perform Principal Component Analysis (PCA) on the feature vectors $ v_f \in \mathbb{R}^{L \times d_h}$, which are constructed by populating the activation probability values of specific neurons while setting irrelevant neurons' positions to zero. As illustrated in Figure~\ref{fig:PCA_qwen_appendix} (Qwen) and Figure~\ref{fig:PCA} (Llama), situational persona neurons exhibit significant variation across topics. 
\dy{Notably, the maximum Euclidean distance between topics within a single domain in the PCA space is comparable to the distance between distinct personality domains. Moreover, nearby points in the PCA space correspond to semantically similar topics, confirming that the observed variation is systematic rather than noise. Given the independence of Big Five personality domains \citep{goldberg2013alternative}, this result indicates that \textbf{distinct situations induces personality-level representational shifts in LLMs}.}

\begin{figure*}[t]
\setlength{\abovecaptionskip}{5pt}   
\setlength{\belowcaptionskip}{0pt}
    \centering
    \includegraphics[width=\linewidth, trim=5pt 0pt 35pt 535pt, clip]{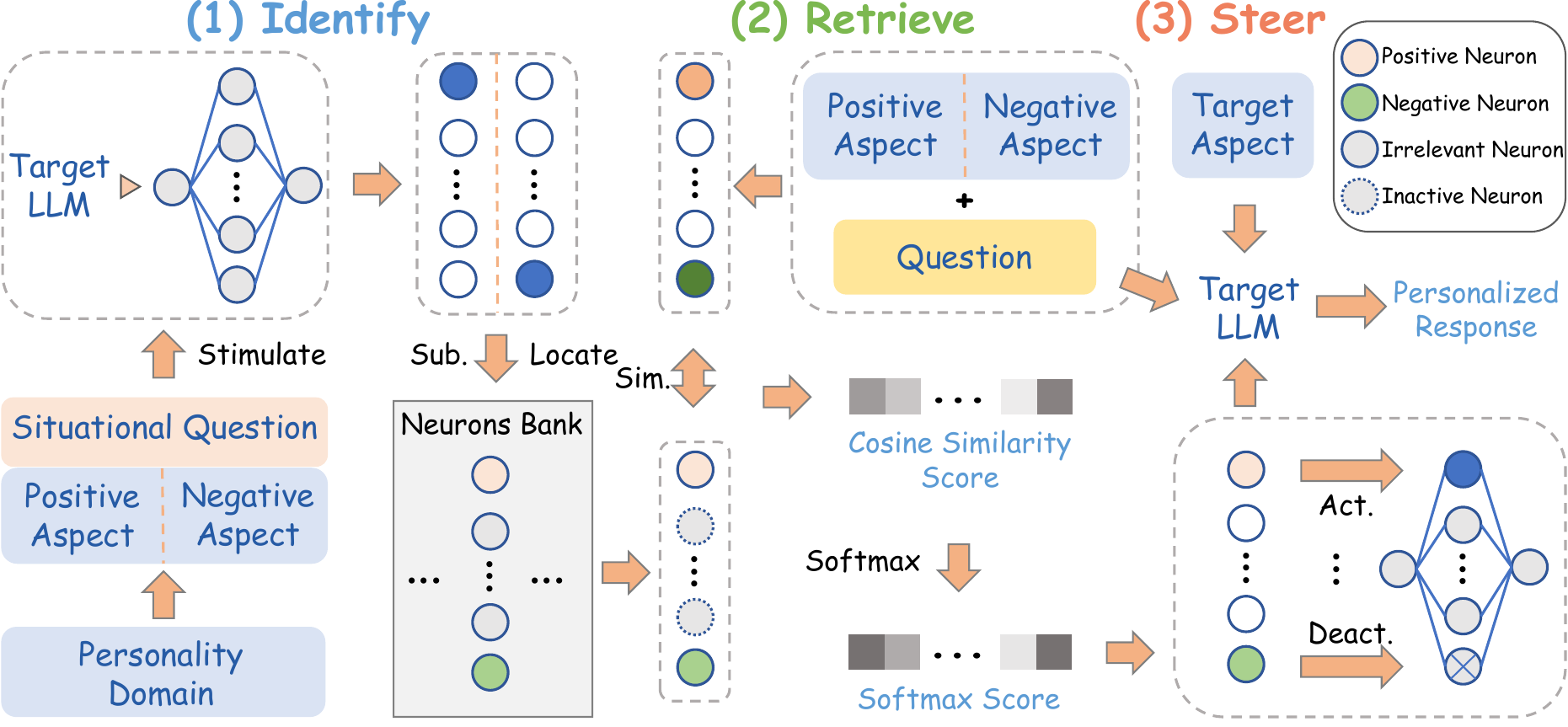} 
    \caption{Overview of the \ourmethod framework, comprising Identification, Retrieval, and Steering phases for accurate personality steering.
    \label{fig:overall_flow_chart}
}
\vspace{-3mm}
\end{figure*}

\paragraph{Layer-Level Perspective} 
We further investigate the variations in the count and proportion of situational personality neurons across diverse topics among all layers. For illustrative clarity, we select eight topics and eight layers spanning the early, middle, and late stages within personality domain \textbf{C}. As shown in Figure~\ref{fig:heatmap}, within a specific personality domain, questions involving different situational topics significantly influence the distribution of neurons controlling LLM personality. Similarly, as illustrated in Figure~\ref{fig:box}, the proportion of situational persona neurons per layer exhibits significant disparity, with a maximum difference of 2.23\% observed in layer 16. Furthermore, the results demonstrate that the middle layers contain the highest concentration of persona neurons across all topics, which aligns with findings from prior studies \citep{skean2025layer, liu-niehues-2025-middle}.

\subsubsection{Stability of Situation-Behavior in LLMs}
\label{sec:Stability of Situation-Behavior}
We next investigate human-like personality patterns of situation-behavior consistency in LLMs. We hypothesize that analogous situations elicit consistent activation patterns in persona neurons, thereby generating coherent personalized behaviors. To investigate this, we select four topics within a specific personality domain, each comprising 1,200 questions covering both aspect. Feature vectors are constructed by aggregating the activation probabilities of situational persona neurons across output tokens, following the methodology in \cref{sec:empiri_situation_impact}. Given the high dimensionality and substantial sample size of vectors, we apply Linear Discriminant Analysis (LDA) following PCA. As illustrated in Figure~\ref{fig:PCA_LDA}, instances from the same topic form distinct and cohesive clusters, suggesting that \textbf{diverse questions within a situation elicit convergent internal activations}. Notably, overlaps between certain clusters reflect semantic correlations across questions, indicating that \textbf{these representations exist within a continuous latent space rather than being strictly binary}.

\section{\ourmethod Framework}

Empirical study in \cref{sec:emp_study} has shown the distinct activation patterns to exert personalized control across different situations and the principle of situation-behavior stability within LLMs. Based on this empirical conclusion, we propose a human-like train-free \textbf{I}dentify\mbox{-}\textbf{R}etr\textbf{i}eve\mbox{-}\textbf{S}teer (\ourmethod) framework to achieve more accurate personality steering. As illustrated in Figure~\ref{fig:overall_flow_chart}, our framework operates in three stages: (1) identifying situational persona neurons (\cref{sec:identify_neurons}); (2) retrieving relevant historical topics via a situation-aware module (\cref{sec:neurons_retrieval}); and (3) modulating the model's personality in novel situation using similarity-weighted steering (\cref{sec:neurons_steering}). To facilitate consistency throughout the paper, we summarize the mathematical notations in \cref{appendix:notation}.

\subsection{Identifying Situational Persona Neurons}
\label{sec:identify_neurons}
Given a personality domain \(\mathcal{D}\), we employ both positive and negative personality prompt to instruct the target LLM to respond to the set of situational questions $\mathcal{Q}_\mathcal{D}^t$ regarding topic $t$. We then calculate the activation probability for all neurons across all output tokens. Specifically, the activation probability of the $i^{th}$ neuron in layer $l$ is computed as:
\begin{equation}\small
\label{eq:locate_neuron}
    P_{i,l}^s = \mathbb{E} \left[ \mathbb{I}\left(\text{act}\left(X^l (W^l_1)_i\right) > 0\right) \mid {\mathcal{D}}^s, t \right]
\end{equation}
Where $\mathbb{I}(\)·\()$  is the indicator function and \(s \in \{+,-\}\) represents the personality aspect. To eliminate impact from irrelevant content and locate the persona neurons, we compute the activation difference across all neurons between positive and negative traits, defined as \(\Delta P = P^+ - P^-\). We classify all neurons, termed \textbf{situational persona neurons}, into positive ($N^{\mathcal{D}^+}_{t}$), negative ($N^{\mathcal{D}^-}_{t}$), or irrelevant ($N^I_{t}$) for specific topic \({t}\) based on whether their activation difference $\Delta P$ is greater than a threshold $10\%$, less than $-10\%$, or falls between these bounds, respectively. For all personality domain and its corresponding topic, we store both $N^{\mathcal{D}^+}_T$ and $N^{\mathcal{D}^-}_T$ with the \textit{location}, \textit{historical activation value}, \textit{raw activation probability of its aspect}, and \textit{activation probability difference} in \textbf{Situational Persona Neurons Bank}, Where \(\mathcal{D} \in \{O,C,E,A,N\} \), and \( T \) is the topic set.

\subsection{Situation-aware Neurons Retrieval}
\label{sec:neurons_retrieval}
We have established that LLMs utilize highly consistent persona neurons to regulate personality expression across similar situations in \cref{sec:Stability of Situation-Behavior}. Consequently, when the model encounters a novel situation, retrieving and adjusting situational persona neurons associated with similar contexts can yield precise personality control.
Specifically, given a novel, topic-agnostic question and a target personality domain $\mathcal{D}$, we firstly compute the activation difference vector $\Delta v$ between the positive and negative aspects (same as the subtraction operation in \cref{sec:identify_neurons}). Secondly, for neuron belonging to $\mathcal{D}$ and specific topic \(t\), we construct the feature vectors $v^{\mathcal{D}}_{t} \in \mathbb{R}^{L \times d_h}$ using the  raw activation probability values across positive and negative neurons, while setting irrelevant neurons' position to zero. We then compute the cosine similarity score between $\Delta v$ with $v^{\mathcal{D}}_{t}$ across all topics in $T$:
\begin{equation}\small
\text{Sim}(v^{\mathcal{D}}_{t}, \Delta \mathbf{v}) =  \frac{v^{\mathcal{D}}_{t} \cdot \Delta \mathbf{v}}{\|v^{\mathcal{D}}_{t}\| \cdot \|\Delta \mathbf{v}\|}   \quad t \in T
\end{equation}
To effectively accentuate the distinctions between scores, we apply a softmax function on all similarity scores, obtaining the normalized situation similarity coefficients $e_t$ for each topic $t$:
\begin{equation}\small
\label{eq:Softmax_coefficients}
e_t = \text{Softmax} \left ( \text{Sim}(v^\mathcal{D}_{t}, \Delta \mathbf{v}) \right ) \quad t \in T
\end{equation}

\subsection{Situational Personality Steering}
\label{sec:neurons_steering}
The soft situation similarity coefficients quantify the semantic alignment between the current situation and reference situations. Motivated by the multi-contextual nuances of questions in $\mathcal{Q}$ identified in \cref{sec:Stability of Situation-Behavior}, we evaluate the precision of these coefficients in Appendix~\ref{appendix:Coefficients_analysis}. The demonstrated high accuracy validates the reliability of this semantic alignment, enabling nuanced, situational personality steering that remains robust even in complex multi-contextual scenarios. Building on this validation, for a target personality aspect $\mathcal{D}$, we amplify the activation of situational persona neurons via coefficient-weighted modulation, ensuring the intervention is proportional to situational relevance:
\begin{equation}\small
\label{eq:edit_positive}
    n' = n_{\text{ori}} +  \sum\nolimits_{t \in T} e_t \cdot \gamma \cdot a_{\text{95}} \cdot f(\Delta P),  \quad  \text{if } n \in N^{\mathcal{D}^+}_t
\end{equation}
Where $n_{\text{ori}}$ represents the original activation, $\gamma$ is to control the magnitude of adjustment, and $a_{95}$ represents the \(95^{th}\) percentile of historical activation value of this neuron to constrain modification. Activation differences are also accounted for by applying a softmax-based smoothing function, defined as \({f(x)} = \frac{1}{1+e^{-10\cdot(|x|-0.15)}}\), to the value. we introduce a threshold $\sigma$ and deactivate those negative situational persona neurons with corresponding similarity coefficients greater than $\sigma$:
\begin{equation}\small
\label{eq:edit_negative}
    n' = \min(0, n_{\text{ori}}), \quad \text{if } n \in \bigcup\nolimits_{t: e_t > \sigma} N^{\mathcal{D}^-}_t
\end{equation}

\begin{table*}[t]
    \setlength{\abovecaptionskip}{3pt}   
    \setlength{\belowcaptionskip}{0pt}
    \centering

    \newcolumntype{L}[1]{>{\raggedright\arraybackslash}p{#1}}
    \newcolumntype{C}[1]{>{\centering\arraybackslash}p{#1}}

    \resizebox{1.0\linewidth}{!}{
        \begin{tabular}{L{2cm}  C{1cm}C{1cm}  C{1cm}C{1cm}  C{1cm}C{1cm}  C{1cm}C{1cm}  C{1cm}C{1cm}  C{1cm}C{1cm}}
        \toprule
        
        \multirow{2}{*}{\bf{Methods}} & 
        \multicolumn{2}{c}{\bf{A}} & 
        \multicolumn{2}{c}{\bf{C}} & 
        \multicolumn{2}{c}{\bf{E}} & 
        \multicolumn{2}{c}{\bf{N}} & 
        \multicolumn{2}{c}{\bf{O}} & 
        \multicolumn{2}{c}{\bf{Average}} \\
        
        \cmidrule(lr){2-3} \cmidrule(lr){4-5} \cmidrule(lr){6-7} \cmidrule(lr){8-9} \cmidrule(lr){10-11} \cmidrule(lr){12-13}
        
         & mean$\uparrow$ & var$\downarrow$ & mean$\uparrow$ & var$\downarrow$ & mean$\uparrow$ & var$\downarrow$ & mean$\uparrow$ & var$\downarrow$ & mean$\uparrow$ & var$\downarrow$ & mean$\uparrow$ & var$\downarrow$ \\
        \midrule
        \grow \bf{\gray{SFT}}    & \gray{9.87} & \gray{0.25} & \gray{9.23} & \gray{0.85} & \gray{9.86} & \gray{0.15} & \gray{9.42} & \gray{0.75} & \gray{9.66} & \gray{0.44} & \gray{9.61} & \gray{0.49} \\
        
        \bf{SP}& \textbf{9.72} & \textbf{0.34} & \underline{9.24} & 1.06 & 9.50 & 1.02 & 7.18 & 1.22 & 6.31 & 1.14 & 8.39 & 0.96 \\
        \textbf{\(P^2\)}        & \underline{9.68} & \underline{0.42} & \underline{9.24} & 1.18 & 9.46 & 0.68 & 9.54 & 0.66 & \underline{9.21} & 1.19 & \underline{9.43} & 0.83 \\
        
        \bf{ActAdd}       & 8.20 & 2.90 & 6.61 & 2.75 & 8.84 & 1.44 & 8.90 & 1.78 & 8.52 & 1.83 & 8.20 & 2.10 \\
        \bf{NPTI}         & 9.64 & 0.49 & \textbf{9.25} & \underline{0.66} & \underline{9.86} & \underline{0.14} & \underline{9.92} & \underline{0.07} & 8.50 & \underline{1.08} & \underline{9.43} & \underline{0.49} \\
        \midrule
        \brow \bf{\ourmethod}    & 9.36 & 0.46 & 9.09 & \textbf{0.55} & \textbf{9.90} & \textbf{0.09} & \textbf{9.94} & \textbf{0.06} & \textbf{9.63} & \textbf{0.36} & \textbf{9.59} & \textbf{0.30} \\
        \bottomrule
        \end{tabular}
    }
    \caption{Personality performance results on \textsc{PersonalityBench} . We report the mean score and variance for each personality domain, along with the overall average. The \textbf{best} and \underline{second-best} results in every column are marked. Our method is \sethlcolor{lightblue}\hl{highlighted}, while SFT results serve as the \sethlcolor{gray90}\hl{upper bound} for reference purposes.}
    \label{table:main_exp}
    \vspace{-1mm}
\end{table*}

\begin{table*}[t]
    \setlength{\abovecaptionskip}{3pt}   
    \setlength{\belowcaptionskip}{0pt}
    \centering

    \newcolumntype{L}[1]{>{\raggedright\arraybackslash}p{#1}}
    \newcolumntype{C}[1]{>{\centering\arraybackslash}p{#1}}

    \resizebox{1.0\linewidth}{!}{
        \begin{tabular}{L{2cm}  C{1cm}C{1cm}  C{1cm}C{1cm}  C{1cm}C{1cm}  C{1cm}C{1cm}  C{1cm}C{1cm}  C{1cm}C{1cm}}
        \toprule
        
        \multirow{2}{*}{\bf{Methods}} & 
        \multicolumn{2}{c}{\bf{A}} & 
        \multicolumn{2}{c}{\bf{C}} & 
        \multicolumn{2}{c}{\bf{E}} & 
        \multicolumn{2}{c}{\bf{N}} & 
        \multicolumn{2}{c}{\bf{O}} & 
        \multicolumn{2}{c}{\bf{Average}} \\
        
        \cmidrule(lr){2-3} \cmidrule(lr){4-5} \cmidrule(lr){6-7} \cmidrule(lr){8-9} \cmidrule(lr){10-11} \cmidrule(lr){12-13}
        
         & mean$\uparrow$ & var$\downarrow$ & mean$\uparrow$ & var$\downarrow$ & mean$\uparrow$ & var$\downarrow$ & mean$\uparrow$ & var$\downarrow$ & mean$\uparrow$ & var$\downarrow$ & mean$\uparrow$ & var$\downarrow$ \\
        \midrule
        \grow \bf{\gray{SFT}}    & \gray{9.67} & \gray{0.50} & \gray{8.83} & \gray{0.84} & \gray{9.23} & \gray{1.03} & \gray{9.17} & \gray{0.95} & \gray{9.47} & \gray{0.57} & \gray{9.27} & \gray{0.78} \\
        
        \bf{SP}& \textbf{9.66} & \textbf{0.59} & \textbf{9.02} & 1.08 & 9.11& 1.42 & 8.42 & 1.16 & 7.29 & 1.67 & 8.70 & 1.18 \\
        \bf{\(P^2\)}        & \underline{9.31} & 1.02 & \underline{8.73} & 1.16 & 9.18 & 1.05 & 8.98 & 1.13 & \underline{9.28} & 0.82 & \underline{9.10} & 1.04 \\
        
        \bf{ActAdd}         & 7.72 & 0.99  & 6.13 & \textbf{0.33}  & 8.06 & 0.82  & 8.13 & 0.77  & 6.93 & 1.57  & 7.39 & 0.90 \\
        \bf{NPTI}         & 8.82 & 0.80 & 8.51 & 0.94 & \underline{9.58} & \underline{0.69} & \underline{9.53} & \textbf{0.45} & 9.04 & \textbf{0.40} & 9.09 & \underline{0.66} \\
        \midrule
        \brow \bf{\ourmethod}    & 9.09 & \underline{0.64} & 8.66 & \underline{0.72} & \textbf{9.63} & \textbf{0.56} & \textbf{9.59} & \underline{0.48} & \textbf{9.33} & \underline{0.48} & \textbf{9.26} & \textbf{0.58} \\
        \bottomrule
        \end{tabular}
    }
    \caption{Personality performance results on SPBench. Notation is consistent with Table~\ref{table:main_exp}.}
    \label{table:balance_main_exp}
    \vspace{-3mm}
\end{table*}

\section{Experiments}
\label{sec:experiments}
\subsection{Experimental Setup}

\noindent \textbf{Benchmarking Dataset}.~ We employ test set from \textsc{PersonalityBench} \citep{deng2024neuronbasedpersonalitytraitinduction} . Notably, these situational questions lack explicit situation labels, whose categories may fall outside the topic set  \( T \). We introduce the \underline{S}ituational \underline{P}ersonality \underline{Bench}mark (\textbf{SPBench}) as a supplementary test set comprising 90 situational questions per personality domain distributed across 30 topic categories, with construction  detail and intrinsic evaluations provided in Appendix~\ref{appendix:Spbench} and ~\ref{appendix:human_evalautino_spbench}. Such settings guarantee the assessment on both \textit{generalization to unseen situational situation}, and \textit{the robustness in comprehensive benchmark}. 
 
\noindent \textbf{Evaluation Metrics}.~ We evaluate personality expression (mean), stability (variance), and response fluency using a 5-point LLM-as-a-Judge scale. Higher scores indicate stronger trait presence or fluency quality. Thus, personality scores for negative aspects are inverted using $(6 - \text{score})$. For each personality domain, we report the results by summing its positive and negative aspect scores.

\noindent \textbf{Baselines}.~ We adopt a diverse range of LLM personalization baselines covering both inference-time and training-time alignment. These include prompt-based methods (Simple Prompt (SP), $P^2$ \citep{10.5555/3666122.3666588}), direct steering methods (ActAdd \citep{turner2024steeringlanguagemodelsactivation}, NPTI \citep{deng2024neuronbasedpersonalitytraitinduction}), and Supervised Fine-tuning (SFT) using LoRA \citep{hu2021loralowrankadaptationlarge}, serving as \textit{upper bound} for reference purpose. Details of implementing these baselines are presented in Appendix~\ref{appendix:baselines}.

\noindent \textbf{Implementation Details}.~  
To ensure a fair comparison, we maintain consistent $\gamma$ values for both \ourmethod and NPTI (1.4 for \textsc{PersonalityBench} and 1.1 for SPBench). LLM-as-a-Judge evaluations are conducted using the \textit{gpt-4o-20240806} \citep{openai2024gpt4ocard} API with greedy decoding. With the exception of \cref{sec:more_models_comparison}, Llama-3-8B-Instruct serves as the target LLM throughout \cref{sec:experiments} and \cref{sec:further_analysis}.


\subsection{Experimental Results}
\noindent \textbf{Results on \textsc{PersonalityBench} }~  As shown in Table~\ref{table:main_exp}, SP achieves optimal performance on A domain, while exhibiting significant deficiencies regarding N and O. These results underscore the inherent instability of prompt-based methods as previously discussed in \cref{sec:introduction}. In contrast, \ourmethod outperforms all baselines in terms of mean and variance of personality performance scores for E, N, O, the overall average across all traits \( \left( mean=9.59, variance=0.30 \right ) \), and score variance on C domain. By surpassing all direct steering baselines, our approach demonstrates more precise neuron control, enabling the model to exhibit personality with greater accuracy and consistency. Furthermore, \ourmethod achieves performance marginally below the SFT baseline while remaining train-free. These superior results demonstrate our framework's resource efficiency, controllability, and generalization to unseen situational situations. Fluency scores are analyzed in Appendix~\ref{appendix:fluency_score}.

\begin{figure*}[t]
    \centering
    \begin{subfigure}[b]{0.49\textwidth} 
        \centering
        \includegraphics[width=\linewidth, trim=20pt 751pt 882pt 260pt, clip]{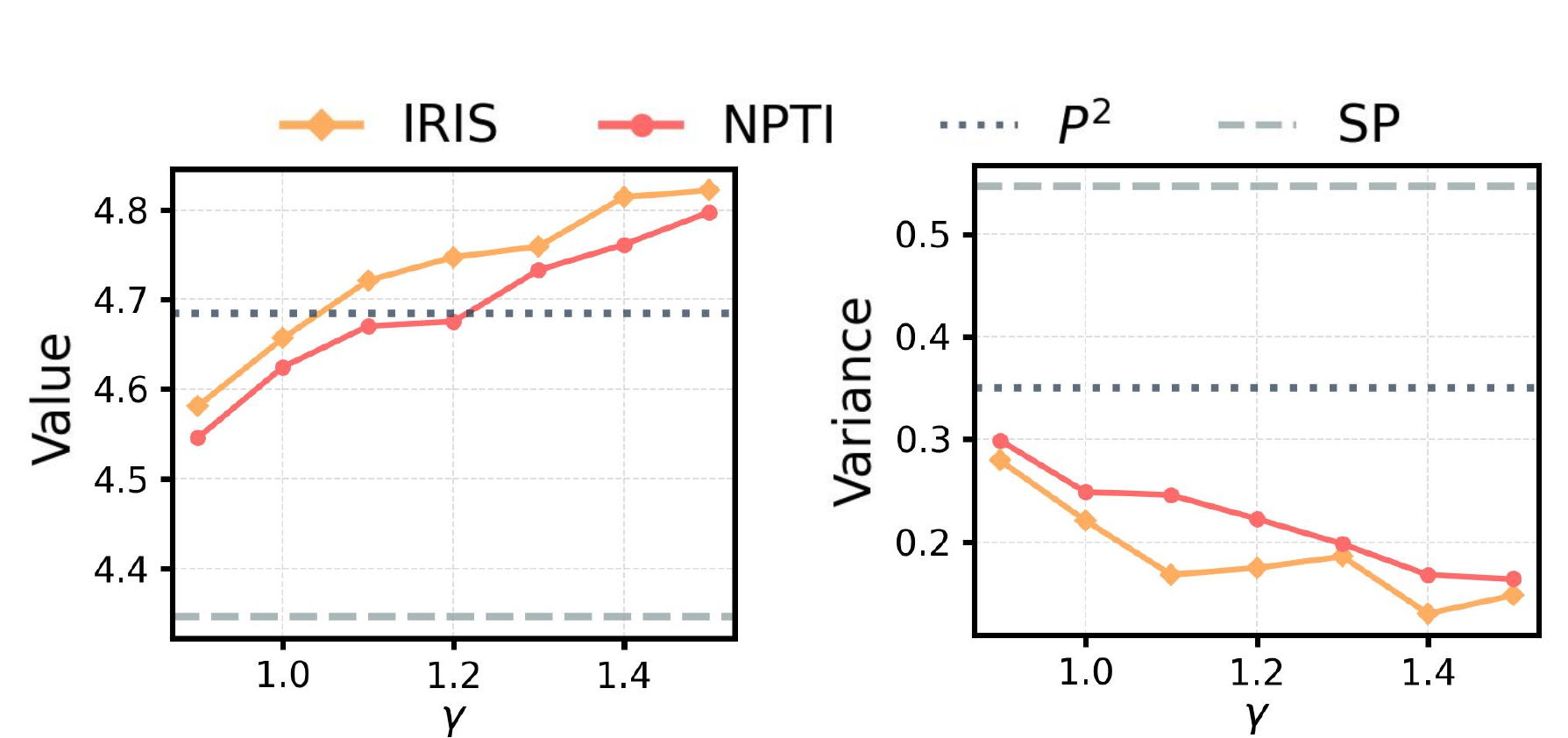}
        \caption{\textsc{PersonalityBench} }
        \label{fig:ablation_gamma_npti}
    \end{subfigure}
    \hfill 
    \begin{subfigure}[b]{0.49\textwidth}
        \centering
        \includegraphics[width=\linewidth, trim=20pt 766pt 882pt 260pt, clip]{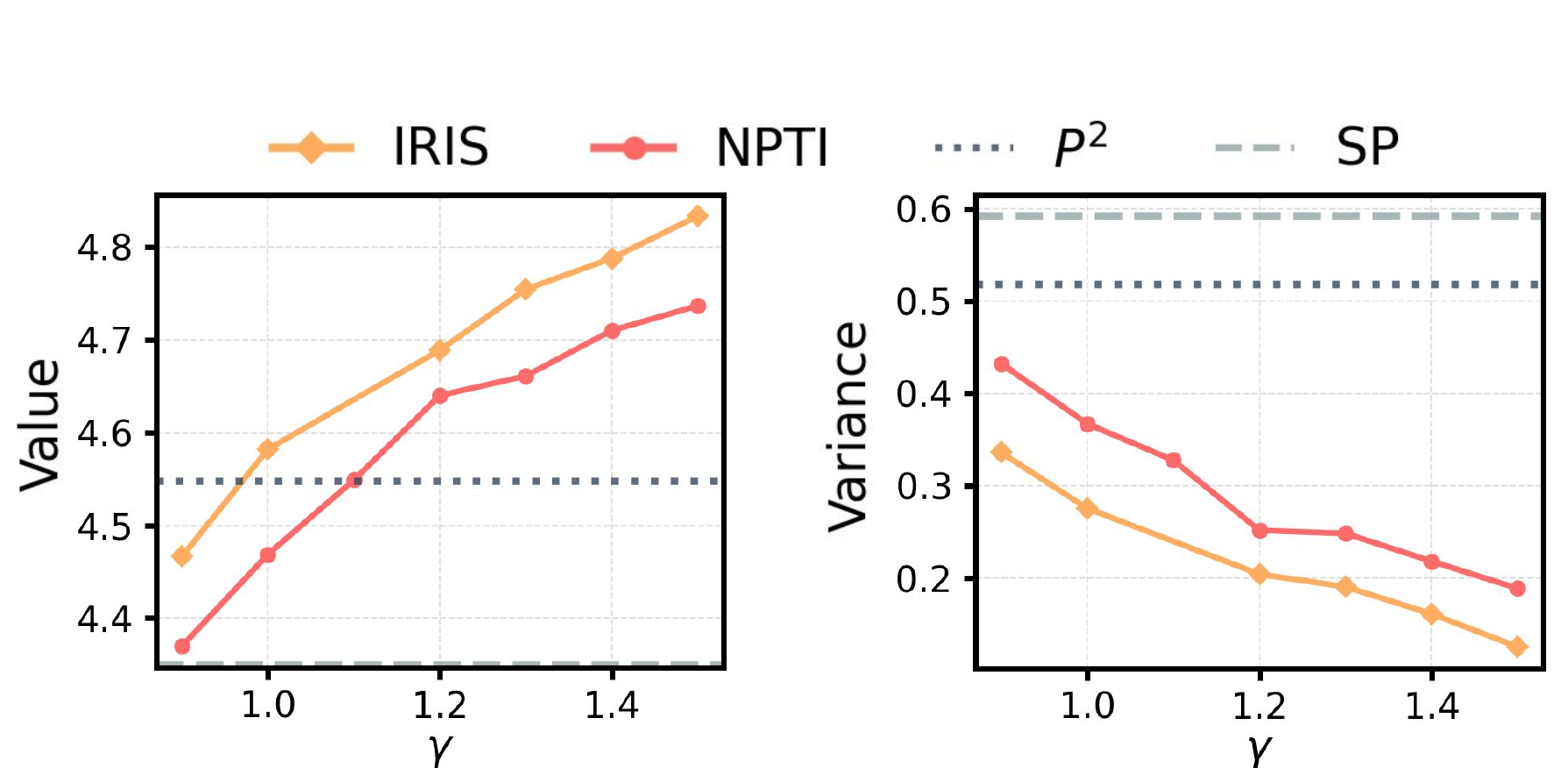}
        \caption{SPBench}
        \label{fig:ablation_gamma_spbench}
    \end{subfigure}
    
    \par\vspace{-0.07cm}
    
    
    \begin{subfigure}[b]{0.49\textwidth} 
        \centering
        \includegraphics[width=\linewidth, trim=0pt 751pt 879pt 261pt, clip]{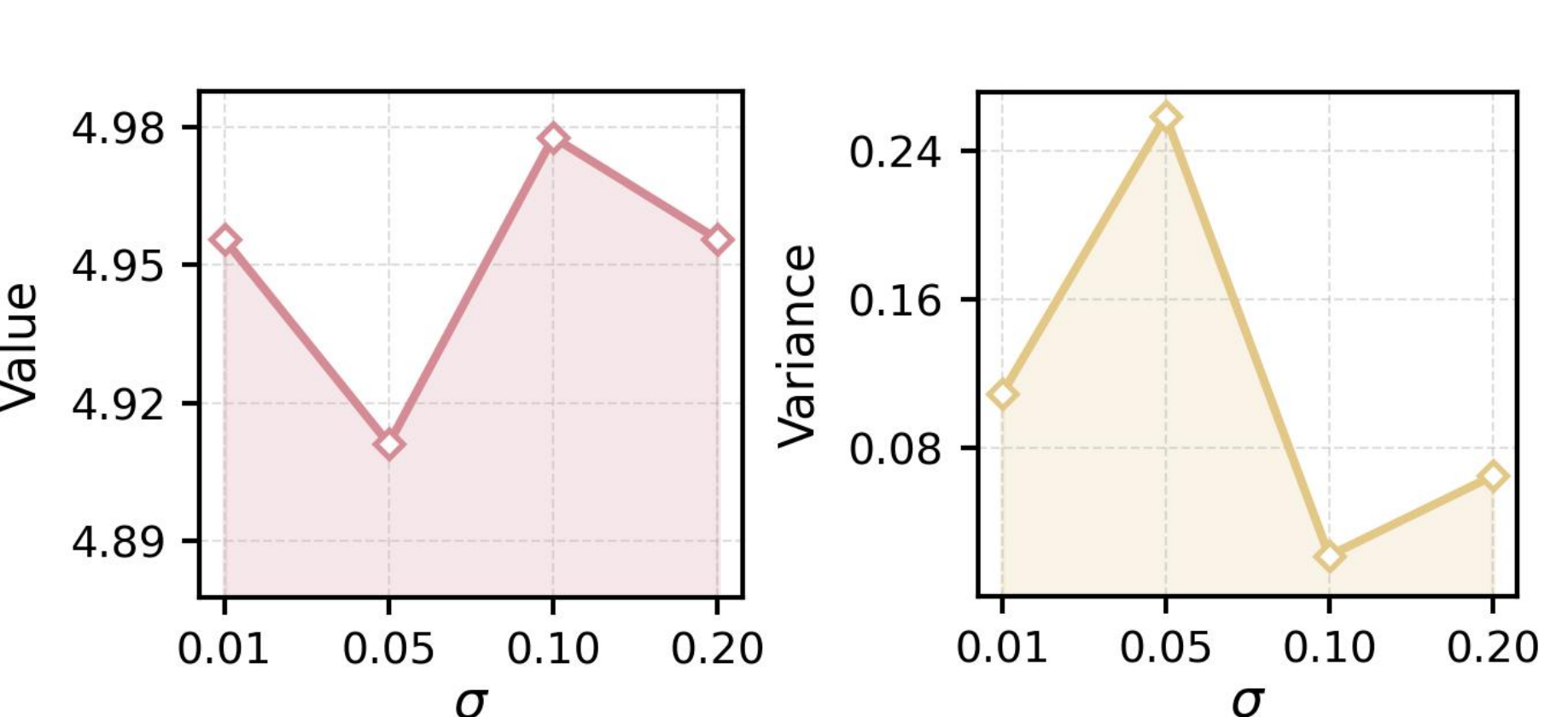}
        \caption{Personality Score}
        \label{fig:ablation_sigma_a}
    \end{subfigure}
    \hfill 
    \begin{subfigure}[b]{0.49\textwidth}
        \centering
        \includegraphics[width=\linewidth, trim=0pt 751pt 879pt 260pt, clip]{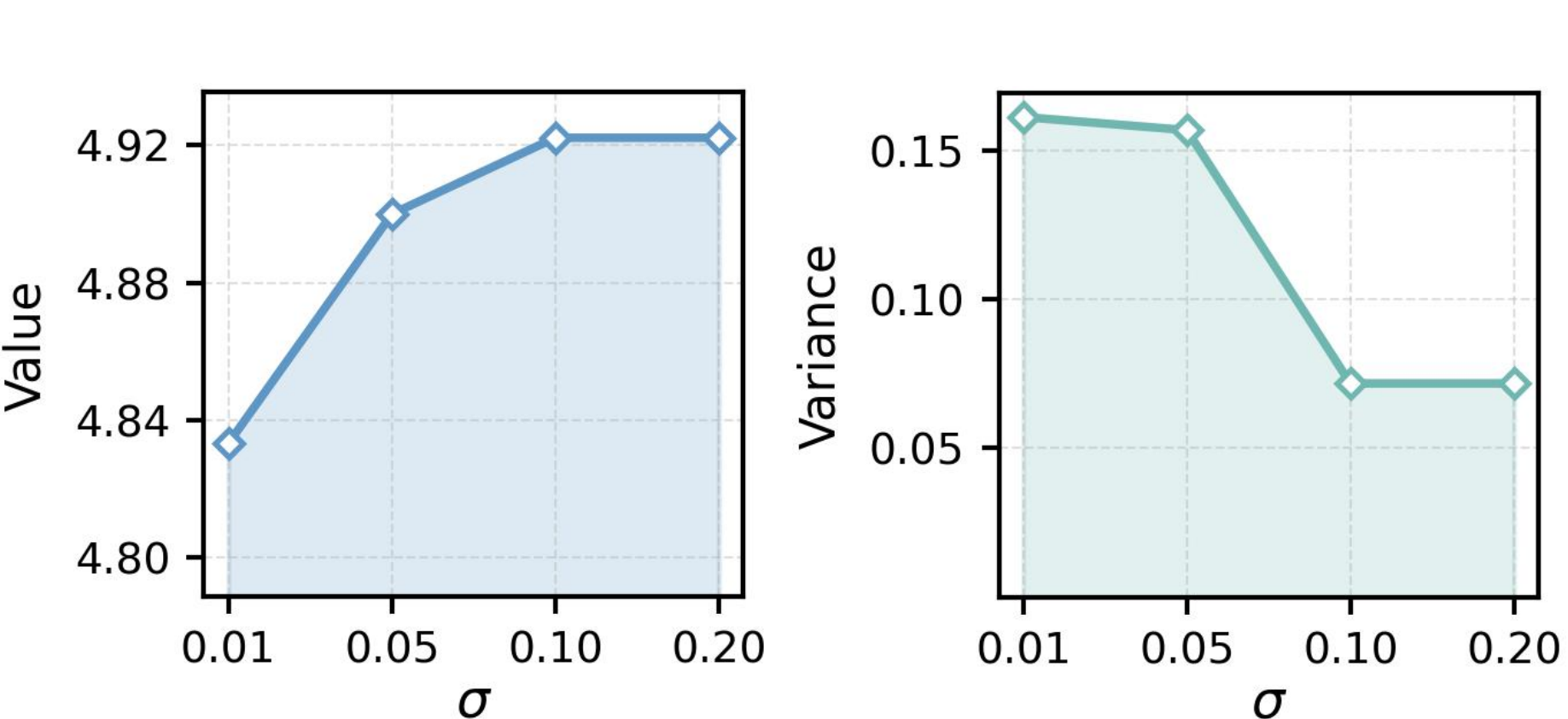}
        \caption{Fluency}
        \label{fig:ablation_sigma_b}
    \end{subfigure}

     \vspace{-0.08cm}
    
    \caption{Ablation study results: (a) and (b) depict the personality assessment performance across two distinct test sets under different \(\gamma\) settings. (c) and (d) illustrate the impact of varying $\sigma$ settings across all evaluation metrics.}
    \label{fig:ablation_sigma}
    \vspace{-3mm}
\end{figure*}
\noindent \textbf{Results on SPBench}~  Notably, \textsc{PersonalityBench} is a relatively simple benchmark where baseline performances saturate near theoretical maximums. While SPBench matches its per-domain question count, it offers a more challenging standard by covering a broader range of situations. As shown in Table~\ref{table:balance_main_exp}, \ourmethod consistently outperforms all baselines on average, underscoring its robustness in navigating complex scenarios. Furthermore, several baselines suffer performance degradation in specific personality domains within SPBench. This suggests that existing methods struggle with personality control in complex situations, further validating SPBench as a comprehensive benchmark.

\section{Further Analysis}
\label{sec:further_analysis}

\begin{table}[t]
    \setlength{\abovecaptionskip}{3pt}   
    \setlength{\belowcaptionskip}{0pt}
    \centering
    \small 
    \setlength{\tabcolsep}{5pt} 

    \setlength{\aboverulesep}{0pt}
    \setlength{\belowrulesep}{0pt}
    \renewcommand{\arraystretch}{1.35}

    \newcolumntype{L}[1]{>{\raggedright\arraybackslash}p{#1}}
    \newcolumntype{C}[1]{>{\centering\arraybackslash}p{#1}}

    \begin{tabular}{L{1cm}  C{0.65cm} C{0.65cm} C{0.65cm} C{0.65cm}  C{1.6cm}}
        \toprule
        
        \multirow{2}{*}{\bf{Method}} & \multicolumn{4}{c}{\bf{Ranking Distribution (\%)}} & \multirow{2}{*}{\bf{Mean Rank} $\downarrow$} \\
        
        \cmidrule{2-5} 
        
        & \bf{1st} & \bf{2nd} & \bf{3rd} & \bf{4th} & \\
        
        \midrule
        
        SP & 16.3 & 22.5 & 28.7 & 32.6 & 2.78 \\
        \(P^2\) & \underline{29.2} & 22.5 & 25.3 & 23.0 & \underline{2.42} \\
        NPTI & 19.1 & 28.7 & 23.0 & 29.2 & 2.62 \\
        
        \midrule
        \bf{\ourmethod} & \textbf{35.4} & 26.4 & 23.0 & 15.2 & \textbf{2.18} \\
        
        \bottomrule
    \end{tabular}
    \caption{Human evaluation results showing the ranking distribution and mean rank of all methods. }
    \label{table:ranking_distribution}
    \vspace{-3mm}
\end{table}

\begin{table*}[t]
    \setlength{\abovecaptionskip}{3pt}   
    \setlength{\belowcaptionskip}{0pt}
    \centering
    \newcolumntype{L}[1]{>{\raggedright\arraybackslash}p{#1}}
    \newcolumntype{C}[1]{>{\centering\arraybackslash}p{#1}}

    \resizebox{1.0\linewidth}{!}{
        \begin{tabular}{L{1.8cm} L{1.8cm}  C{1cm}C{1cm}   C{1cm}C{1cm}   C{1cm}C{1cm}  C{1cm}C{1cm}   C{1cm}C{1cm}  C{1cm}C{1cm}}
        \toprule
        
        \multirow{2}{*}{\bf{Model}} & \multirow{2}{*}{\bf{Methods}} & 
        \multicolumn{2}{c}{\bf{A}} & 
        \multicolumn{2}{c}{\bf{C}} & 
        \multicolumn{2}{c}{\bf{E}} & 
        \multicolumn{2}{c}{\bf{N}} & 
        \multicolumn{2}{c}{\bf{O}} & 
        \multicolumn{2}{c}{\bf{Average}} \\
        
        \cmidrule(lr){3-4} \cmidrule(lr){5-6} \cmidrule(lr){7-8} \cmidrule(lr){9-10} \cmidrule(lr){11-12} \cmidrule(lr){13-14}
        
         & & mean$\uparrow$ & var$\downarrow$ & mean$\uparrow$ & var$\downarrow$ & mean$\uparrow$ & var$\downarrow$ & mean$\uparrow$ & var$\downarrow$ & mean$\uparrow$ & var$\downarrow$ & mean$\uparrow$ & var$\downarrow$ \\
        \midrule
        
        \multirow{4}{*}{\bf{Qwen}} 
        & \bf{SP}& 9.41 & 0.58 & 9.11 & 0.88 & 9.46 & 0.66 & 8.74 & 0.98 & 6.41 & 1.68 & 8.63 & 0.95 \\
        & \textbf{\(P^2\)}  & 9.46 & 0.54 & 9.28 & 0.57 & 9.39 & 0.77 & 9.45 & 0.90 & 9.18 & 0.79 & 9.35 & 0.71 \\
        & \bf{NPTI}         & \underline{9.76} & \underline{0.23} & \textbf{9.82} & \underline{0.16} & \textbf{9.96} & \textbf{0.04} & \underline{9.96} & \underline{0.04} & \underline{9.64} &\underline{0.23} & \underline{9.83} & \underline{0.14} \\
        & \bf{\ourmethod} & \textbf{9.94} & \textbf{0.05} & \underline{9.80} & \textbf{0.17} & \underline{9.93} & \underline{0.08} & \textbf{10.00} & \textbf{0.00} & \textbf{9.74} & \textbf{0.38} & \textbf{9.89} & \textbf{0.10} \\
        
        \midrule
        
        \multirow{4}{*}{\bf{Gemma}} 
        & \bf{SP}& \bf{9.96} & \bf{0.04} & \underline{9.07} & \underline{0.66} & 9.63 & 0.43 & 8.95 & 0.79 & 8.39 & 1.62 & 9.20 & 0.71 \\
        & \textbf{\(P^2\)}  & 9.76 & 0.21 & 8.61 & 0.91 & 9.61 & 0.48 & 9.55 & 0.58 & 9.32 & 0.48 & 9.37 & 0.53 \\
        & \bf{NPTI}   & 9.62 & 0.24 & 8.90 & 1.56 & \underline{9.91} & \underline{0.17} & \bf{9.99} & \bf{0.01} & \underline{9.34} & \bf{0.25} & \underline{9.55} & \underline{0.44} \\
        & \bf{\ourmethod} & \underline{9.92} & \underline{0.07} & \bf{9.75} & \bf{0.26} & \bf{9.96} & \bf{0.04} & \underline{9.98} & \underline{0.04} & \bf{9.70} & \underline{0.28} & \bf{9.86} & \bf{0.14} \\
        
        \bottomrule
        \end{tabular}
    }
    \caption{Personality performance results on Qwen3-8B and gemma-3-12b-it, with  consistent Notation in Table~\ref{table:main_exp}.}
    \label{table:difference_model_comparison}
    \vspace{-3mm}
\end{table*}

\begin{table*}[t]
    \setlength{\abovecaptionskip}{3pt}   
    \setlength{\belowcaptionskip}{0pt}
    \centering
    \newcolumntype{L}[1]{>{\raggedright\arraybackslash}p{#1}}
    \newcolumntype{C}[1]{>{\centering\arraybackslash}p{#1}}

    \resizebox{1.0\linewidth}{!}{
        \begin{tabular}{L{1.8cm}  C{1cm}C{1cm}   C{1cm}C{1cm}   C{1cm}C{1cm}  C{1cm}C{1cm}   C{1cm}C{1cm}  C{1cm}C{1cm}}
        \toprule
        
        \multirow{2}{*}{\bf{Methods}} & 
        \multicolumn{2}{c}{\bf{A}} & 
        \multicolumn{2}{c}{\bf{C}} & 
        \multicolumn{2}{c}{\bf{E}} & 
        \multicolumn{2}{c}{\bf{N}} & 
        \multicolumn{2}{c}{\bf{O}} & 
        \multicolumn{2}{c}{\bf{Average}} \\
        
        \cmidrule(lr){2-3} \cmidrule(lr){4-5} \cmidrule(lr){6-7} \cmidrule(lr){8-9} \cmidrule(lr){10-11} \cmidrule(lr){12-13}
        
         & mean$\uparrow$ & var$\downarrow$ & mean$\uparrow$ & var$\downarrow$ & mean$\uparrow$ & var$\downarrow$ & mean$\uparrow$ & var$\downarrow$ & mean$\uparrow$ & var$\downarrow$ & mean$\uparrow$ & var$\downarrow$ \\
        
        \hline \hline
        \multicolumn{13}{c}{\textbf{Qwen3-0.6B}} \\
        \hline
        \bf{SP}           & 6.73 & 2.21 & 6.78 & 0.89 & 6.83 & 1.30 & 7.73 & 1.00 & 6.30 & 2.19 & 6.87 & 1.50 \\
        \textbf{\(P^2\)}  & 7.57 & 1.90 & 7.42 & 1.11 & 8.73 & 0.79 & 8.18 & 1.07 & 7.60 & 1.60 & 7.90 & 1.29 \\
        \bf{NPTI}         & 9.03 & 0.46 & 8.74 & \bf{0.59} & 9.56 & 0.31 & 9.87 & 0.10 & 9.18 & 0.35 & 9.28 & 0.36 \\
        \bf{IRIS}         & \bf{9.53} & \bf{0.31} & \bf{8.90} & 0.70 & \bf{9.83} & \bf{0.10} & \bf{10.00} & \bf{0.00} & \bf{9.73} & \bf{0.14} & \bf{9.60} & \bf{0.25} \\
        
        \hline \hline
        \multicolumn{13}{c}{\textbf{Qwen3-8B}} \\
        \hline
        \bf{SP}           & 9.41 & 0.58 & 9.11 & 0.88 & 9.46 & 0.66 & 8.74 & 0.98 & 6.41 & 1.68 & 8.63 & 0.95 \\
        \textbf{\(P^2\)}  & 9.46 & 0.54 & 9.28 & 0.57 & 9.39 & 0.77 & 9.45 & 0.90 & 9.18 & 0.79 & 9.35 & 0.71 \\
        \bf{NPTI}         & 9.76 & 0.23 & \bf{9.82} & \bf{0.16} & \bf{9.96} & \bf{0.04} & 9.96 & 0.04 & 9.64 & \bf{0.23} & 9.83 & 0.14 \\
        \bf{IRIS}         & \bf{9.94} & \bf{0.05} & 9.80 & 0.17 & 9.93 & 0.08 & \bf{10.00} & \bf{0.00} & \bf{9.74} & 0.38 & \bf{9.89} & \bf{0.10} \\
        
        \hline \hline
        \multicolumn{13}{c}{\textbf{Qwen3-14B}} \\
        \hline
        \bf{SP}           & 9.47 & 0.24 & 9.13 & 0.53 & 9.21 & 0.43 & 8.90 & 0.74 & 7.36 & 2.23 & 8.81 & 0.83 \\
        \textbf{\(P^2\)}  & 9.64 & 0.19 & 8.63 & 0.78 & 9.43 & 0.30 & 9.59 & 0.31 & 9.07 & 0.58 & 9.27 & 0.43 \\
        \bf{NPTI}         & 9.84 & 0.07 & 9.70 & 0.13 & 9.86 & 0.09 & \bf{10.00} & \bf{0.00} & 9.64 & \bf{0.15} & 9.81 & 0.09 \\
        \bf{IRIS}         & \bf{9.96} & \bf{0.02} & \bf{9.84} & \bf{0.08} & \bf{9.97} & \bf{0.03} & \bf{10.00} & \bf{0.00} & \bf{9.71} & \bf{0.15} & \bf{9.90} & \bf{0.06} \\
        
        \bottomrule
        \end{tabular}
    }
    \caption{Personality performance results on the Qwen3 series across different model scales.}
    \label{table:difference_model_size_comparison}
    \vspace{-3mm}
\end{table*}

\subsection{Human Evaluation}  
We recruit five human evaluators to rank the responses from four top-performing methods. We conduct a pre-evaluation training for all human evaluators and verify their inter-annotator agreement, with more details provided in Appendix \ref{appendix:human_evaluation}. We aggregate the evaluation results to calculate both the distribution of rankings across all methods and the overall average ranks, as presented in Table \ref{table:ranking_distribution}. The results demonstrate that \ourmethod achieves the highest proportion of first-place rankings and the most favorable average rank. These findings are consistent with the outcomes of the LLM-as-a-Judge evaluation, further validating the effectiveness of our proposed framework.

\subsection{Effect of Hyperparameters}

\noindent \textbf{Impact of \(\gamma\)}~ The parameter $\gamma$ in Eq.~\ref{eq:edit_positive} regulates the steering magnitude of activation values for positive persona neurons. We investigate its impact on overall personality performance scores by conducting an ablation study with values ranging from 0.9 to 1.5 across two test sets. As illustrated in Figure~\ref{fig:ablation_gamma_npti} and Figure~\ref{fig:ablation_gamma_spbench}, both \ourmethod and NPTI underperform compared to $P^2$ at lower $\gamma$ values, primarily constrained by their preset prompt. However, performance improves for both methods as $\gamma$ increases. Notably, our method consistently outperforms NPTI. For instance, our approach at $\gamma = 1.3$ surpasses NPTI at $\gamma = 1.4$. These results validate that our framework outperforms baselines by leveraging precise neuron modification to achieve superior efficiency and accuracy in personality elicitation under identical hyperparameter settings.

\noindent \textbf{Impact of \(\sigma\)}~ The parameter $\sigma$ in Eq.~\ref{eq:edit_negative} governs the selection threshold for negative neurons. A smaller $\sigma$ implies a larger number of negative neurons are selected for deactivation. To investigate this, we conducted an ablation study on $\sigma$ on the \textsc{PersonalityBench} , while keeping other hyper-parameters fixed. We evaluated four candidate values: $\{0.01, 0.05, 0.10, 0.20\}$, analyzing their impact on both personality performance and fluency of results on positive aspect of A domain. As shown in Figure~\ref{fig:ablation_sigma_a}, both the personality performance score and variance exhibit an initial increase followed by a decline as $\sigma$ increases. Conversely, Figure~\ref{fig:ablation_sigma_b} illustrates the results for response fluency: as $\sigma$ increases, fewer negative neurons are deactivated, leading to improved generation fluency. Consequently, we set $\sigma = 0.10$ given the stable fluency between 0.10 and 0.20.

\subsection{Generalizability Across Different Models}
\label{sec:more_models_comparison}
To evaluate the robustness of our method across different LLMs, we extend our experiments to include two additional advanced models: Qwen3-8B and gemma-3-12B-it \citep{gemmateam2025gemma3technicalreport}. These models introduce architectural diversity. Specifically, Qwen utilizes the SiLU \citep{elfwing2017sigmoidweightedlinearunitsneural} activation function, consistent with our primary model Llama-3-8B-Instruct, while Gemma employs GeLU \citep{hendrycks2023gaussianerrorlinearunits}. We maintain constant hyperparameters across all settings, with the exception of $\gamma$, which is adjusted to $1.0$ for Qwen and $1.3$ for Gemma. As illustrated in Figure~\ref{table:difference_model_comparison}, \ourmethod ranks first in the majority of personality domains and second in the remainder, achieving the best overall performance on both models. These results demonstrate the robustness of our approach across varying model architectures.

\subsection{Impact of Model Scaling on Generalizability}
To investigate the scalability and consistency of our method across varying model sizes, we conduct a scaling analysis utilizing different scales within the Qwen3 family, specifically Qwen3-0.6B, Qwen3-8B, and Qwen3-14B. All hyperparameters are kept strictly constant across these models. As presented in Table~\ref{table:difference_model_size_comparison}, existing methods exhibit noticeable scaling inconsistencies, where increasing model parameters does not guarantee proportional performance gains. Notably, scaling up the model size can occasionally result in performance degradation. For instance, baseline methods (e.g., $P^2$ and NPTI) on the 8B model unexpectedly outperform their 14B counterparts in the C domain. In contrast, \ourmethod demonstrates highly robust scalability. As the model parameters increase, our method not only steadily improves in average performance but also significantly reduces variance (from 0.25 on 0.6B to 0.06 on 14B). Furthermore, \ourmethod consistently maintains SOTA overall performance across all three scales. These findings strongly validate the reliability and effectiveness of our approach, proving its capability to harness the power of larger models stably without suffering from prompt sensitivity.

\subsection{Inherent Situational Impact on Personality Expression}
Situational factors inherently influence the expression of personality. To investigate this nuanced impact, we collect results from \ourmethod on SPBench and calculate a standardized score for each topic. Specifically, the raw scores for the positive and negative aspects are independently standardized by subtracting their respective means and dividing by their standard deviations. These standardized values are then averaged to derive the final score for each topic. This metric quantifies the directional impact of a topic, where higher values indicate a positive deviation. Table \ref{table:Inher_Impact_of_topics} shows the topics exhibiting the most significant deviations across all domains. Notably, "History and Nostalgia" yields a strongly negative score in Neuroticism, suggesting that factual topics inherently suppress Neuroticism by eliciting calmer responses, thereby attenuating the effect of the intended persona adjustment. 
\dy{Qualitative case studies are provided in Appendix~\ref{appendix:case_study} to explicitly reveal the distinct characteristics of situational impact on the responses steered by \ourmethod.}

\begin{table}[t]
    \setlength{\abovecaptionskip}{3pt}   
    \setlength{\belowcaptionskip}{0pt}
    \centering
    \small 
    \setlength{\tabcolsep}{10pt} 

    \setlength{\aboverulesep}{0pt}
    \setlength{\belowrulesep}{0pt}
    \renewcommand{\arraystretch}{1.35} 

    \newcolumntype{C}[1]{>{\centering\arraybackslash}p{#1}}

    \begin{tabular}{C{0.5cm} C{2.6cm}  C{1.5cm}}
        \toprule
        
        \bf{Domain} & \bf{Topic} & \bf{Std. Score} \\ 
        \midrule
        
        \multirow{2}{*}{A} & Fashion\_Style & 1.038 \\
                           & Creativity\_Inspiration & -1.318 \\
        \midrule
        \multirow{2}{*}{C} & Relationships\_Dating & 0.783 \\
                           & Music\_Entertainment & -0.995 \\
        \midrule
        \multirow{2}{*}{E} & History\_Nostalgia & 0.946 \\
                           & Philosophy\_Ethics & -0.840 \\
        \midrule
        \multirow{2}{*}{N} & Pop\_Culture\_trends & 0.750 \\
                           & History\_Nostalgia & -1.636 \\
        \midrule
        \multirow{2}{*}{O} & Family\_Parenting & 0.739 \\
                           & Education\_Learning & -0.369 \\
        
        \bottomrule
    \end{tabular}
    \caption{Results of Situational Impact on Personality.}
    \label{table:Inher_Impact_of_topics}
    \vspace{-3mm}
\end{table}

\section{Related Works}
Early training-based LLM personalization approaches rely on large-scale data to align models with general human preferences \citep{longpre2023flancollectiondesigningdata, li2024personalizedlanguagemodelingpersonalized}. While subsequent research addresses granularity limitations by leveraging fine-grained datasets \citep{bu-etal-2025-alignx} or training user-specific LoRA modules \citep{tan-etal-2024-democratizing, wang-etal-2025-beyond-profile}, these methods incur significant computational overhead, necessitate complex data curation, and lack the flexibility to adapt to evolving user preferences. Alternatively, training-free strategies employ contextual guidance, ranging from explicit prompting \citep{tan2024phantompersonabasedpromptingeffect, zhu2025personality} to implicit mechanisms such as Retrieval-Augmented Generation for accessing personalized content \citep{richardson2023integratingsummarizationretrievalenhanced, salemi2024optimizationmethodspersonalizinglarge}. Although flexible, these approaches are often unstable due to their sensitivity to provided context quality. Another training-free paradigm involves direct internal steering, such as editing attention heads \citep{zhu2025personality}, perturbing hidden states among layers via probing classifiers \citep{ju2025probing}, or modifying personality-related neuron activations \citep{deng2024neuronbasedpersonalitytraitinduction}. However, these interventions typically lack robust psychological grounding, treating personality as a static attribute, which fail to capture the dynamic nature of human behavior across varying situations. 

\section{Conclusion}

In this paper, we introduce \ourmethod, a neuron-based situational personality steering framework in LLMs. Grounded in our validation of situational dependency and consistent situation\mbox{-}behavior patterns within LLMs, \ourmethod integrates psychological theory with mechanistic intervention to achieve precise and human-like personality steering. Extensive experiments demonstrate that \ourmethod significantly outperforms existing paradigms on two comprehensive benchmarks and exhibits robust generalization across different model architectures. Furthermore, our in-depth analysis provides valuable insights to guide future research in this domain.

\section*{Acknowledgments}
This work is supported by Anhui Province Key Research and Development Plan (202304a05020045), National Natural Science Foundation of China under Grant 62406098, the Singapore Ministry of Education (MOE)
Academic Research Fund (AcRF) Tier 1 grant (Proposal ID: 24-SIS-SMU-002), and the National Research Foundation Singapore under the AI Singapore Programme (AISG Award No: AISG3-RPGV-2025-016).

\section*{Limitations}
We discuss the limitations from two perspectives:
\paragraph{Trade-offs Between Personalization and Instruction Following Capabilities}
Our case studies reveal that steering models toward specific personality traits can occasionally compromise instruction following, \dy{which is a common issue in LLM personalization approaches \citep{handa2025personalityprobellmevaluation,dearaujo2025persistentpersonasroleplayinginstruction}}. For instance, inducing the negative aspect of Openness may cause the model to refuse interaction, reflecting realistic "closed" human behaviors. This highlights an inherent trade-off between personality expression and task completion. Given the diverse applications of LLM personalization, \textbf{this balance requires scenario-specific consideration}. We also conduct comprehensive discussion on this issue, including evaluations of response fluency (Appendix~\ref{appendix:fluency_score}) and general instruction-following capabilities (Appendix~\ref{appendix:general_capability}).

\paragraph{Granularity of Situation Topics}
A second limitation lies in the granularity of our retrieval mechanism, which currently relies on a taxonomy of 30 historical reference topics \dy{from \cite{ding-etal-2023-enhancing}}. The fixed segmentation may be insufficient for certain nuances, \textbf{underscoring the need for more open-ended situational modeling}. Nevertheless, our framework effectively retrieves multiple relevant situations based on current unseen situational question. Furthermore, the superior performance across two benchmarks, and in-depth analysis in \cref{sec:neurons_steering} and \cref{appendix:Coefficients_analysis} demonstrate \ourmethod 's efficiency and robustness to unseen complex situations.

\section*{Ethical Considerations}
We strictly adhere to the licensing terms of the models utilized in this study. Llama-3-8B-Instruct is governed by the Meta Llama 3 Community License Agreement\footnote{https://huggingface.co/meta-llama/Meta-Llama-3-8B/blob/main/LICENSE}, Qwen3-8B is distributed under the Apache License Version 2.0\footnote{https://huggingface.co/Qwen/Qwen3-8B/blob/main/LICENSE}, and gemma-3-12b-it is subject to the Gemma Terms of Use\footnote{https://ai.google.dev/gemma/terms}. Additionally, the Roberta-large employed for the verification experiments in \cref{appendix:Coefficients_analysis} operates under the MIT License. We have properly cited all software, datasets, and models used, and we affirm that all presented results are intended exclusively for scientific research purposes. Our proposed benchmarking dataset SPBench consists of synthetic LLM-generated questions. It contains no personally identifiable information, such as names or individual identifiers, nor does it include offensive content.
\bibliography{main}

\newpage

\appendix

\section*{Appendix}

\label{sec:appendix}

\section{Situational Topics Set}
\label{appendix:topic_set}
Figure~\ref{label:complete_topic_set} presents the complete situational topics set used in this paper defined in  UltraChat \citep{ding-etal-2023-enhancing}. This set comprises 30 comprehensive categories covering various aspects of daily life, facilitating the investigation of personality manifestations and mechanisms across diverse situations.

\section{Notations}
\label{appendix:notation}
\begin{table}[htbp]
    \centering
    \small 
    \setlength{\tabcolsep}{8pt}
    \setlength{\aboverulesep}{0pt}
    \setlength{\belowrulesep}{0pt}
    \renewcommand{\arraystretch}{1.5} 

    \newcolumntype{L}[1]{>{\centering\arraybackslash\bfseries}m{#1}} 
    
    \newcolumntype{P}[1]{>{\raggedright\arraybackslash}m{#1}}

    \begin{tabular}{L{1.1cm} || P{5.4cm}}
        \toprule
        \multirow{1}{*}{Notation} & \multirow{1}{*}{\bf{Description}} \\
        \midrule
        
        $L$ & Number of Layers (Transformer Blocks) in Model \\
        \midrule
        
        $l$ & Specific Layer Index $l$ \\
        \midrule
        
        $X^l$ & Input of Layer $l$ \\
        \midrule
        
        $d$ & Model Dimension \\
        \midrule
        
        $d_h$ & Hidden Dimension of the FFN \\

        \midrule
        
        $\mathcal{Q}$ & Neuron Identification Question Set \\
        
        \midrule
        
        $T$ & Situational Topics Set \\

        \midrule
        
        $t$ & Specific Topic \\
        
        \midrule
        
        $M$ & Total Number of Situational Topics\\
        \midrule
        
        $\mathcal{D}$ & A Specific Personality Domain \\
        \midrule
        
        $P$ & Activation Probability of Neurons \\
        \midrule

        $s$ & Aspects (+, -) of Personality Domain \\
        \midrule

        $e_t$ & Normalized Situation Similarity Coefficients for Topic $t$ \\
        \midrule
        
        $N^{\mathcal{D}^+}_t$ & Situational Persona Neuron Set for Positive Aspects of Domain $\mathcal{D}$ and Topic $t$ \\

        \midrule
        
        $N^{\mathcal{D}^-}_t$ & Situational Persona Neuron Set for Negative Aspects of Domain $\mathcal{D}$ and Topic $t$ \\
        
        \midrule
        
        $N^{I}_t$ & Irrelevant Situational Persona Neuron Set for Topic $t$ \\

        \midrule
        
        $n'$ & Edited Neurons \\
        
        \midrule
        $n$ & Neurons \\
        \bottomrule
    \end{tabular}
    \caption{Summary of notations used in this paper.}
    \label{table:notations}
\end{table}

\begin{table*}[h]
    \centering
    \renewcommand{\arraystretch}{1}
    \setlength{\tabcolsep}{2pt}
    
    \begin{tabular}{l l l}
    \toprule
        Art and culture & Health and wellness & Pop culture and trends \\
        Beauty and self-care & History and nostalgia & Relationships and dating \\
        Creativity and inspiration & Literature and writing & Science and innovation \\
        Diversity and inclusion & Mindfulness and meditation & Social media and communication \\
        Education and learning & Money and finance & Spirituality and faith \\
        Entrepreneurship and business & Music and entertainment & Sports and fitness \\
        Family and parenting & Nature and the environment & Technology \\
        Fashion and style & Personal growth and development & Travel and adventure \\
        Food and drink & Philosophy and ethics & Travel and culture exchange \\
        Gaming and technology & Politics and current events & Work and career \\
    \bottomrule
    \end{tabular}
    \caption{The situational topic set used in this paper are listed in alphabetical order.}
    \label{label:complete_topic_set}
\end{table*}

\section{Comprehensive Empirical Study}
\label{appendix:complete_empirical_study}
\subsection{PCA Analysis of Persona neurons} 
\begin{figure}[htbp]
\setlength{\abovecaptionskip}{5pt}   
\setlength{\belowcaptionskip}{0pt}
    \centering
    \includegraphics[width=\linewidth, trim=5pt 5pt 889pt 768pt, clip]{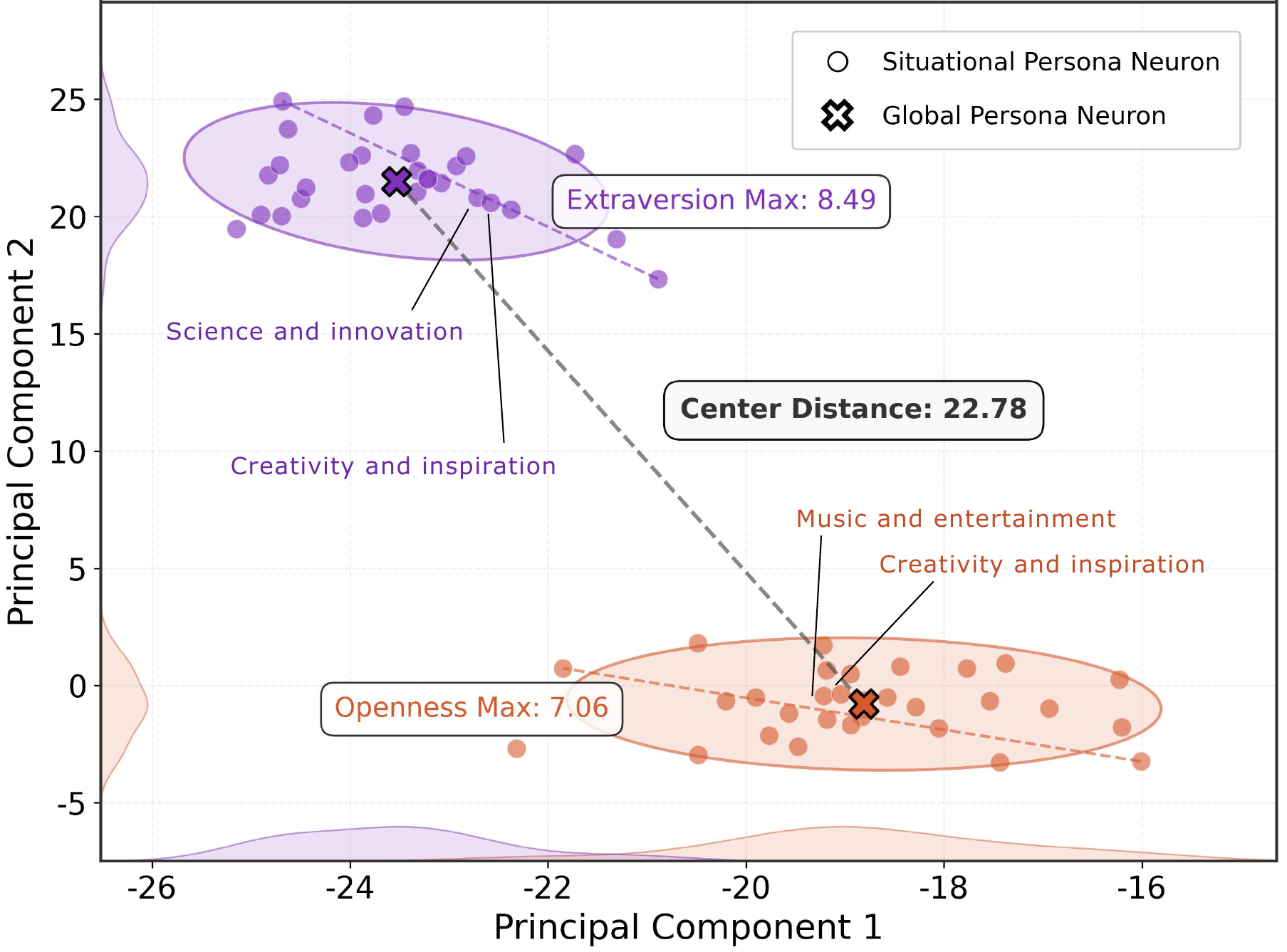} 
    \caption{PCA of situation-dependent and global persona neurons in Llama-3-8B-Instruct, with proximate topic names annotated.
    \label{fig:PCA}
}
\vspace{-3mm}
\end{figure}

\begin{figure*}[h]
    \centering
    \begin{subfigure}[b]{0.49\textwidth} 
        \centering
        \includegraphics[width=\linewidth, trim=0pt 0pt 5pt 80pt, clip]{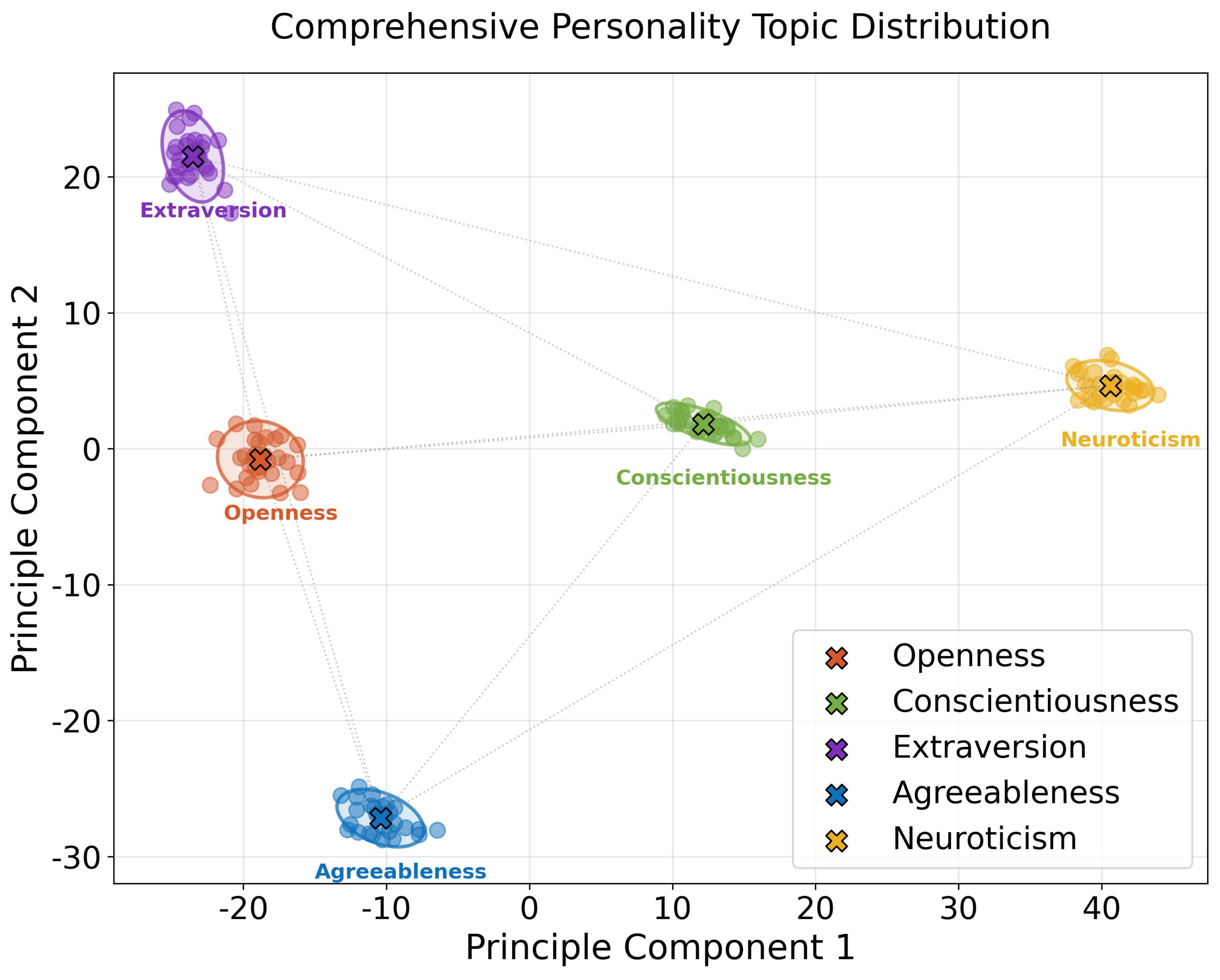}
        \caption{}
        \label{fig:all_personalities_PCA_llama}
    \end{subfigure}
    \hfill 
    \begin{subfigure}[b]{0.49\textwidth}
        \centering
        \includegraphics[width=\linewidth, trim=0pt 5pt 5pt 80pt, clip]{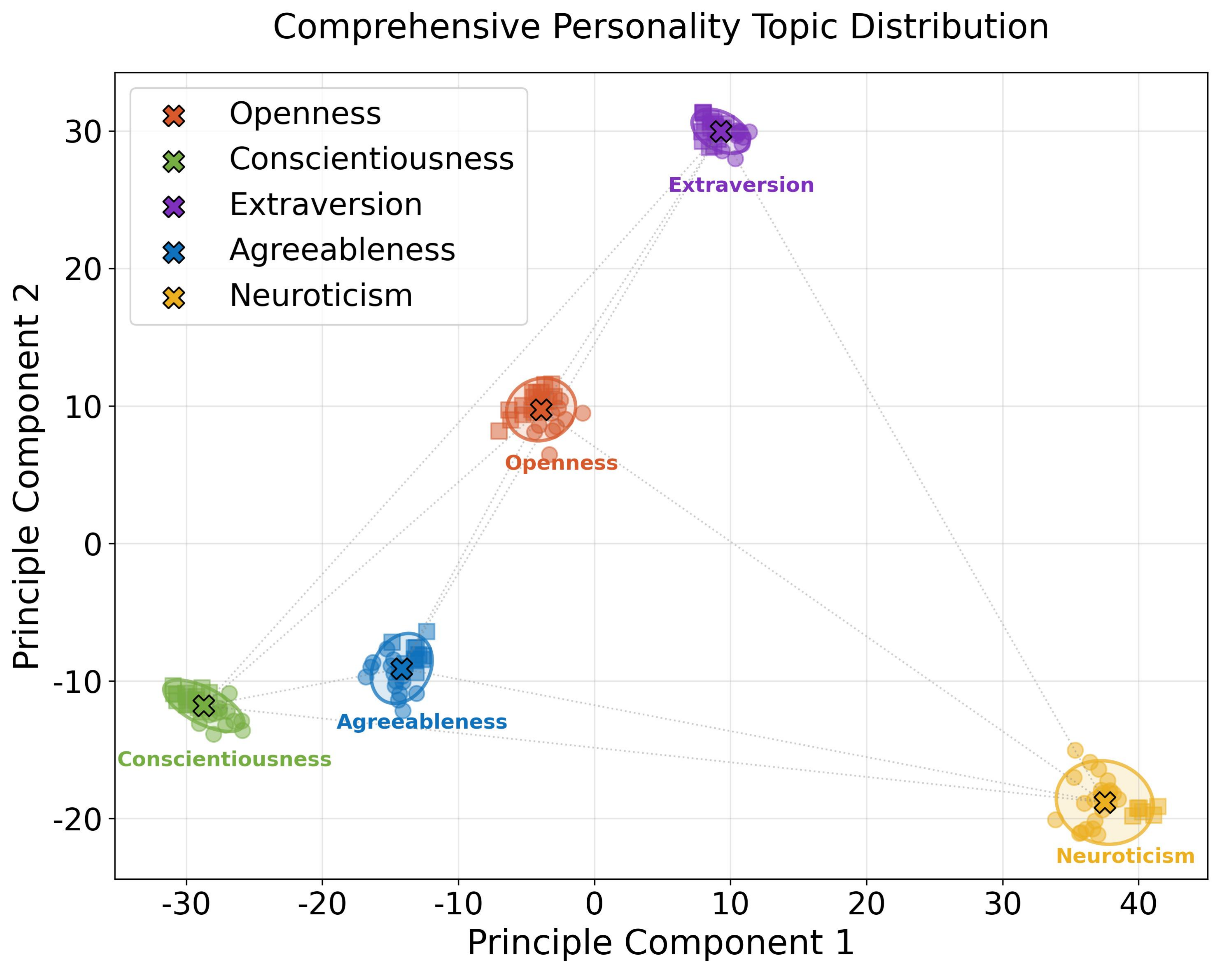}
        \caption{}
        \label{fig:all_personalities_PCA_qwen}
    \end{subfigure}
    
    \caption{Overall PCA results across five personality domains of Llama-3-8B-Instruct (a) and Qwen3-8B (b).}
    \label{fig:all_personalities_PCA}
\end{figure*}

We report the complete results of PCA on the activation patterns of situational and global persona neurons. To demonstrate the universality of our findings across different models, we selected Llama-3-8B-Instruct and Qwen3-8B as our target LLMs. As illustrated in Figure \ref{fig:all_personalities_PCA}, situational persona neurons within the same personality domain exhibit significant variance. Notably, the maximum PCA distance between these neurons within certain domain is of the same order of magnitude as the distance between distinct, highly independent personalities. Furthermore, we mark two most prominent personality domains in Qwen model for detailed analysis, as shown in Figure~\ref{fig:PCA_qwen_appendix}. Following the format of Figure \ref{fig:PCA}, we annotate two proximal topic points within these personality for reference.

\subsection{Layer Analysis of Persona Neurons}  
\begin{figure*}[h]
    \centering
    \begin{subfigure}[b]{0.49\textwidth} 
        \centering
        \includegraphics[width=\linewidth, trim=0pt 0pt 0pt 0pt, clip]{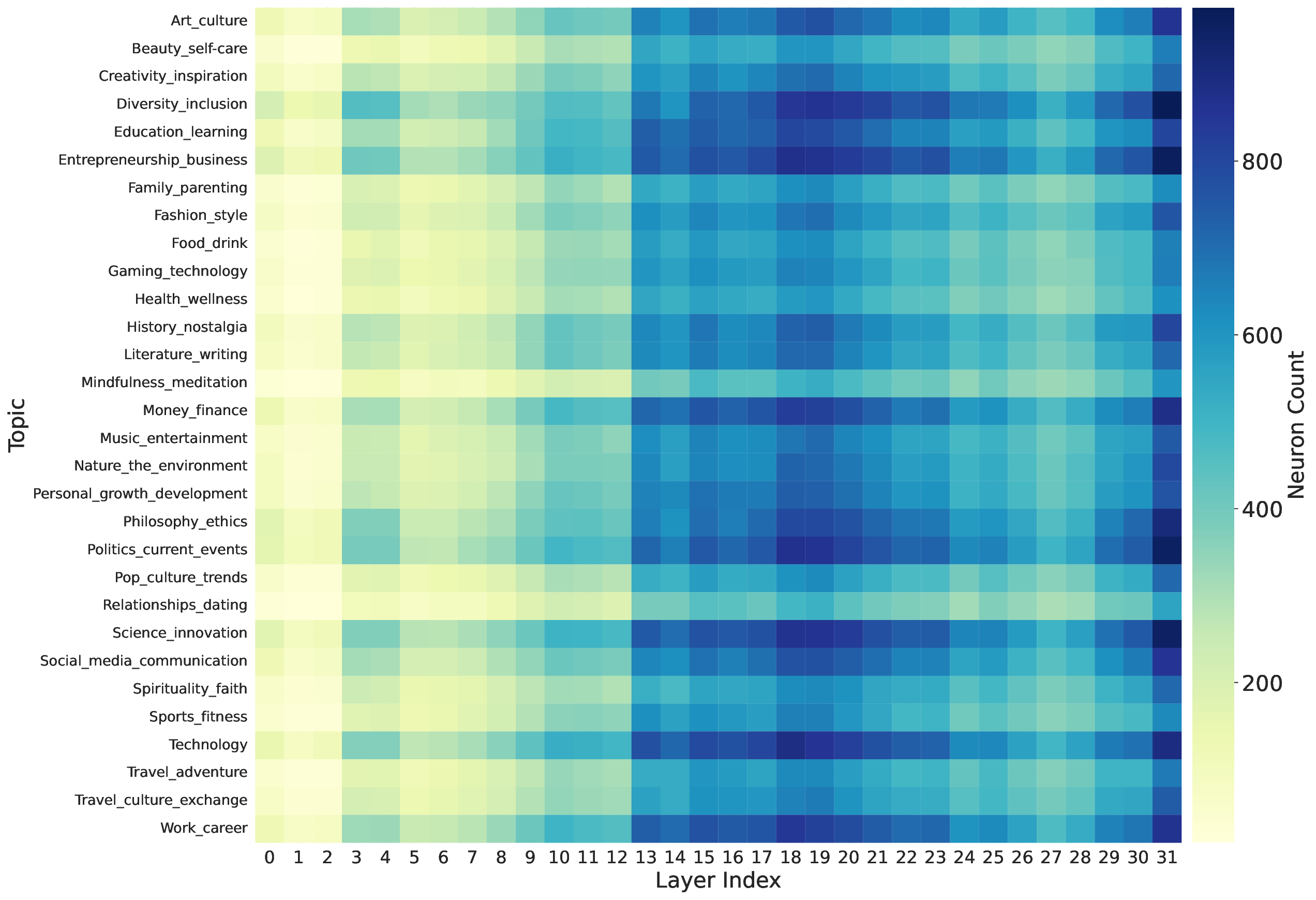}
        \caption{Llama-3-8B-Instruct (32 Layers)}
        \label{fig:overall_heatmap_llama}
    \end{subfigure}
    \hfill 
    \begin{subfigure}[b]{0.49\textwidth}
        \centering
        \includegraphics[width=\linewidth, trim=0pt 0pt 0pt 0pt, clip]{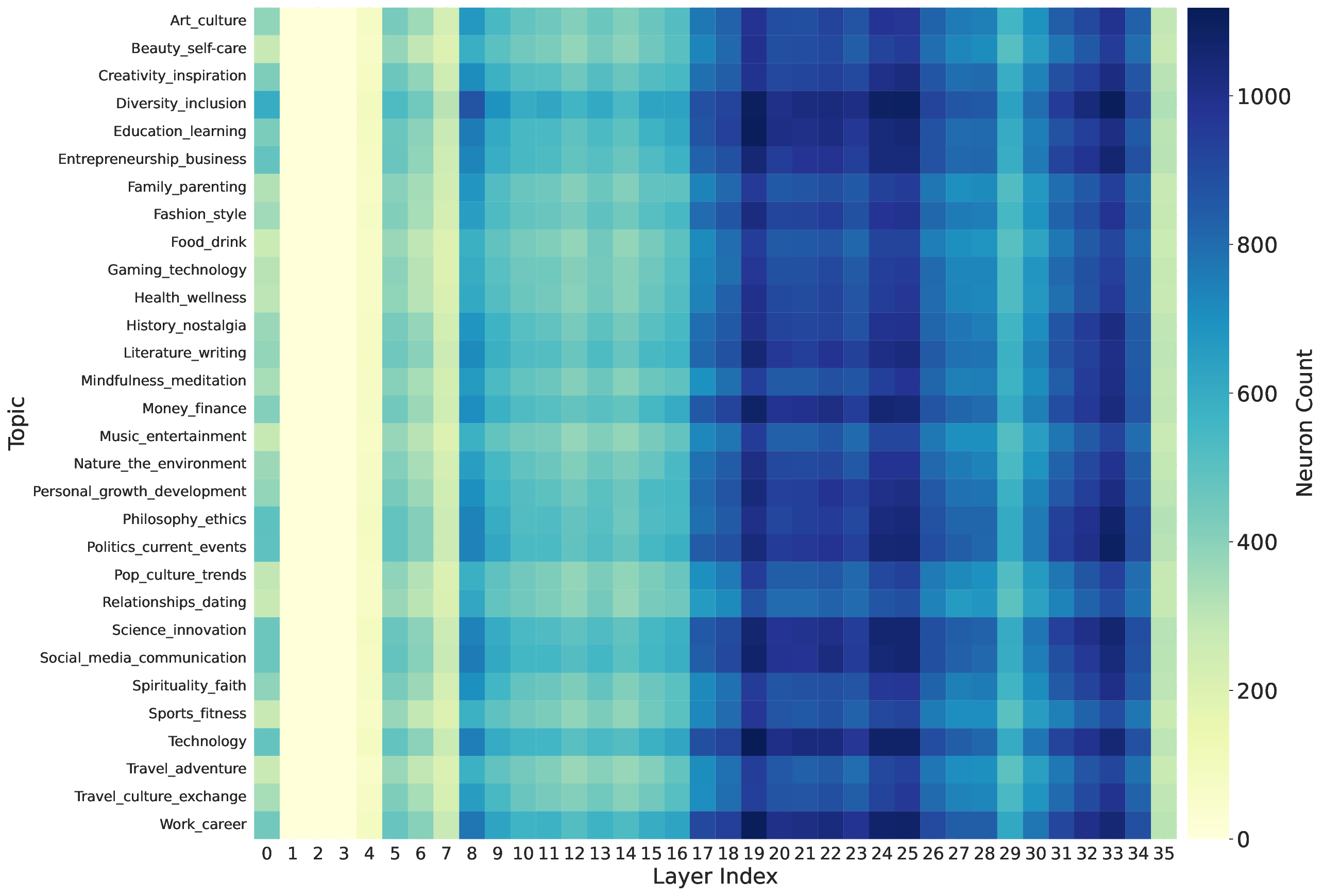}
        \caption{Qwen3-8B (36 Layers)}
        \label{fig:overall_heatmap_qwen}
    \end{subfigure}
    
    \caption{Comprehensive layer analysis of the count distribution of situational persona neurons.}
    \label{fig:overall_heatmap}
\end{figure*}

We present the comprehensive analysis of the situational persona neurons across all layers and diverse topics, with the authentic counts and proportions illustrated in Figure \ref{fig:overall_heatmap} and Figure \ref{fig:overall_box_llama}, respectively. These results characterize the variability of persona neurons associated with traits of specific personality domain when subjected to different situations. Furthermore, both figures clearly illustrate the distribution of these neurons across the models, revealing that the neurons governing personality expression are most densely concentrated in the middle and final layers of the models.


\section{Implementation Details of Baselines}
\label{appendix:baselines}
As for the baselines of personalization methods within LLMs, we choose the followings studied in our experiments. All experiments are conducted on a server equipped with 4$\times$ NVIDIA L40S GPUs.

\noindent \textbf{Simple Prompt:} This method explicitly defines personality by adapting trait descriptors directly into the system prompt (e.g., "You are an extraverted person"). This instructs the model to align its generated response with the target trait.

\noindent \textbf{$P^2$} \citep{10.5555/3666122.3666588}: This approach leverages comprehensive descriptions of personality traits to provide richer semantic constraints. $P^2$ thus induces more consistent and verifiable personified behaviors across diverse situational queries and open-ended tasks.


\noindent \textbf{ActAdd} \citep{turner2024steeringlanguagemodelsactivation}: ActAdd is an inference-time intervention that applies vector addition to the Transformer’s residual stream to steer high-level attributes, specifically personality traits in our experiments. Following authentic methodology, we adopt grid search to determine the optimal hyperparameters for each personality domain. The prompts used to calculate the steering vector are consistent with those employed during the Identify stage in \ourmethod.

\noindent  \textbf{Supervised Fine-Tuning:} This baseline enables the model to internalize target personality traits at the parameter level, typically yielding the most robust and stable behavioral expression. Since it directly modifies model weights rather than relying on prompting or inference-time intervention, this method serves as the "training-based upper bound" for reference purpose. Specifically, we collect all responses generated during the Identify stage as ground truth data. To ensure specialized adaptation, we train a distinct LoRA \citep{hu2021loralowrankadaptationlarge} adapter for each personality aspect, targeting all linear modules with a standard cross-entropy loss. The model is trained using BF16 precision with a maximum sequence length of 4096. Optimization hyperparameters include a learning rate of $1 \times 10^{-4}$, a LoRA rank of 8 ($\alpha=16$), and a per-device batch size of 8. Training for each personality aspect requires approximately 3 GPU hours.

\begin{table}[h]
    \setlength{\abovecaptionskip}{3pt}   
    \setlength{\belowcaptionskip}{0pt}
    \centering
    \small 
    \setlength{\tabcolsep}{4pt} 
    \setlength{\aboverulesep}{0pt}
    \setlength{\belowrulesep}{0pt}
    \renewcommand{\arraystretch}{1.35} 

    \newcolumntype{C}[1]{>{\centering\arraybackslash}p{#1}}
    \newcolumntype{L}[1]{>{\raggedright\arraybackslash}p{#1}}

    \begin{tabular}{L{2cm} C{0.8cm} C{0.8cm} C{0.8cm} C{0.8cm} C{0.8cm}}
        \toprule
        
        \bf{Metric} & \bf{A} & \bf{C} & \bf{E} & \bf{N} & \bf{O} \\ 
        \midrule
        
        \multicolumn{6}{c}{{\textbf{SPBench}}} \\
        \midrule
        

        Top-2 Acc. $\uparrow$     & 13.9 & 18.3 & 20.0 & 12.8 & 27.2  \\
        
        Top-5 Acc. $\uparrow$     & 32.8 & 35.0 & 38.3 & 24.4 & 50.6  \\
        
        Top-8 Acc. $\uparrow$     & 47.2 & 48.9 & 48.9 & 37.8 & 63.3  \\
        \hline \hline
        \multicolumn{6}{c}{\textbf{\textsc{PersonalityBench}}} \\
        \midrule
        

        Top-2 Acc. $\uparrow$     & 6.1 & 16.3 & 13.3 & 20.5 & 20.6  \\
        
        Top-5 Acc. $\uparrow$     & 17.2 & 38.2 & 35.0 & 39.5 & 37.2  \\
        
        Top-8 Acc. $\uparrow$     & 35.6 & 53.4 & 52.8 & 48.5 & 49.4  \\
        \bottomrule
        
    \end{tabular}
    \caption{Accuracy (\%) of soft situation similarity coefficients on SPBench and \textsc{PersonalityBench}, demonstrating high accuracy of Retrieve phase in \ourmethod.}
    \label{table:accuracy_across_traits}
    \vspace{-3mm}
\end{table}

\section{Accuracy Analysis for Soft Situation Similarity Coefficients}
\label{appendix:Coefficients_analysis}
To establish a semantic ground truth, we fine-tune RoBERTa-Large\footnote{https://huggingface.co/FacebookAI/xlm-roberta-large} (325M) \citep{liu2019robertarobustlyoptimizedbert} for each personality domain using all questions paired with single-label situational topics in  $\mathcal{Q}$, employing a 30-dimensional MLP classifier. The resulting softmax probability distributions serve as the reference standard for evaluating the Top-2, Top-5, and Top-8 accuracy of our soft situation similarity coefficients. Our approach demonstrates robust performance across all domains in this 30-class classification task. As shown in Table~\ref{table:accuracy_across_traits}, for the Openness domain, we achieve a Top-2 accuracy of 27.2\% and a Top-8 accuracy of 63.3\%. Considering the inherent misalignment between RoBERTa's and neuron-level semantic space, these results are sufficient to validate that the soft situation similarity coefficients retrieved in \cref{sec:neurons_retrieval} effectively capture the granular situational semantics embedded in the questions. Notably, our method achieves this in a completely training-free manner.

\section{Construction Details of SPBench}
\label{appendix:Spbench}
We follow the methodology to generate situational questions as \textsc{PersonalityBench} \citep{deng2024neuronbasedpersonalitytraitinduction}. For a given personality domain and situational topic, we randomly select a descriptive facet as defined by the Five-Factor Model \citep{mccrae1992introduction}. Example facets for each personality domain are presented in Table~\ref{table:facet_definitions}. Utilizing these facets, we instruct GPT-4o \citep{openai2024gpt4ocard} to generate situational questions via one-shot prompting. The specific prompt template is detailed in the "Generation" section of Figure~\ref{tab:SPBench_prompt_generation_refinement}. Following the initial generation, We employ GPT-4o to further polish the questions, as prompt template described in the "Refinement" section of Figure~\ref{tab:SPBench_prompt_generation_refinement}. We generate three situational questions per topic within each personality domain, resulting in a total of 450 questions (90 questions across 5 personality domains). These questions are designed to be neutral and applicable to both positive and negative personality domains. By incorporating a diverse and balanced array of situational topics, SPBench serves as a more comprehensive benchmark for  evaluation.

\section{Evaluation Details} 

\subsection{Intrinsic Evaluation of SPBench}
\label{appendix:human_evalautino_spbench}
To validate the quality of SPBench, we conduct a comprehensive intrinsic evaluation focusing on two key metrics: Inducement Validity (\textbf{IV}) and Situational Authenticity (\textbf{SA}), both scored on a scale of 1 to 10. IV measures the efficacy of a situational question in activating a target personality trait, thereby prompting the model to exhibit that trait in its response. SA evaluates the realism of the scenario description, ensuring alignment with real-world plausibility. We employ both LLM-as-a-Judge approach and human evaluation for comprehensive assessment. For the automatic evaluation, to mitigate potential self-preference bias from the source model, we utilize the \textit{claude-3-5-sonnet-20241022} API with greedy decoding. For the human evaluation, we hire five graduate students specializing in psychology. Prior to assessment, evaluators undergo rigorous training with manually verified examples to ensure alignment with metric definitions. The criteria provided to human evaluators are identical to those used in the automatic evaluation. Each evaluator assesses situational questions ($5$ personality domains $\times 90$ items), and their scores are averaged to produce the final reported results. The evaluation was conducted over two days, with seven-hour daily sessions (08:00 to 17:00, excluding breaks). Evaluators were compensated at a rate of \$0.30 per question. Results are shown in Table~\ref{table:intrinsic_spbench}, SPBench demonstrates superior quality across both metrics. Average scores for each personality domain are consistently high, with IV reaching 8.33 (automatic) and 8.85 (human), and SA averaging approximately 8.0. These results underscore the robustness and reliability of SPBench as a comprehensive benchmark for personality evaluation.

\begin{table}[htbp]
    \setlength{\abovecaptionskip}{3pt}   
    \setlength{\belowcaptionskip}{0pt}
    \centering
    \small 
    \setlength{\tabcolsep}{4pt} 
    \setlength{\aboverulesep}{0pt}
    \setlength{\belowrulesep}{0pt}
    \renewcommand{\arraystretch}{1.35} 

    \newcolumntype{C}[1]{>{\centering\arraybackslash}p{#1}}
    \newcolumntype{L}[1]{>{\raggedright\arraybackslash}p{#1}}

    \begin{tabular}{L{1cm} C{1cm} C{1cm} C{1cm} C{1cm} C{1cm}}
        \toprule
        
        \bf{Metric} & \bf{A} & \bf{C} & \bf{E} & \bf{N} & \bf{O} \\ 
        \midrule

        \multicolumn{6}{c}{\textbf{Automatic Evaluation}} \\
        \midrule
        \textbf{IV}     & 8.33 & 8.74 & 8.44 & 7.96 & 8.20  \\
        
        \textbf{SA}     & 8.04 & 8.18 & 7.98 & 7.68 & 7.42  \\
        
        \hline \hline
        \multicolumn{6}{c}{\textbf{Human Evaluation}} \\
        \midrule
        

        \textbf{IV}    & 9.10 & 8.98 & 8.52 & 8.92 & 8.71  \\
        
        \textbf{SA}     & 7.83 & 8.57 & 8.16 & 7.97 & 7.89  \\
        \bottomrule
        
    \end{tabular}
    \caption{Intrinsic Evaluation of the SPBench. Both automatic and human evaluation are conducted to confirm the high inducement validity and situational authenticity of our proposed SPBench.}
    \label{table:intrinsic_spbench}
    \vspace{-3mm}
\end{table}

\begin{table*}[htbp]
    \centering
    \newcolumntype{L}[1]{>{\raggedright\arraybackslash}p{#1}}
    \newcolumntype{C}[1]{>{\centering\arraybackslash}p{#1}}

    \resizebox{1.0\linewidth}{!}{
        \begin{tabular}{L{2cm}  C{1cm}C{1cm}  C{1cm}C{1cm}  C{1cm}C{1cm}  C{1cm}C{1cm}  C{1cm}C{1cm}  C{1cm}C{1cm}}
        \toprule
        
        \multirow{2}{*}{\bf{Methods}} & 
        \multicolumn{2}{c}{\bf{A}} & 
        \multicolumn{2}{c}{\bf{C}} & 
        \multicolumn{2}{c}{\bf{E}} & 
        \multicolumn{2}{c}{\bf{N}} & 
        \multicolumn{2}{c}{\bf{O}} & 
        \multicolumn{2}{c}{\bf{Average}} \\
        
        \cmidrule(lr){2-3} \cmidrule(lr){4-5} \cmidrule(lr){6-7} \cmidrule(lr){8-9} \cmidrule(lr){10-11} \cmidrule(lr){12-13}
        
         & mean$\uparrow$ & var$\downarrow$ & mean$\uparrow$ & var$\downarrow$ & mean$\uparrow$ & var$\downarrow$ & mean$\uparrow$ & var$\downarrow$ & mean$\uparrow$ & var$\downarrow$ & mean$\uparrow$ & var$\downarrow$ \\
        \midrule
        
        \grow \bf{\gray{SFT}}    & \gray{9.76} & \gray{0.25} & \gray{9.80} & \gray{0.18} & \gray{9.97} & \gray{0.03} & \gray{9.95} & \gray{0.05} & \gray{9.72} & \gray{0.23} & \gray{9.84} & \gray{0.14} \\
        
        \bf{SP}& 9.77 & 0.27 &9.92 & 0.07 & 10.0 & 0.00 & 10.0 & 0.00 & 9.99 & 0.01 & 9.94 & 0.07 \\
        \bf{\(P^2\)}        & 9.81 & 0.42 & 9.91 & 0.08 & 10.0 & 0.00 & 10.0 & 0.00 & 9.87 & 0.18 & 9.92 & 0.09 \\
        
        \bf{ActAdd}       & 8.69 & 1.58 & 8.92 & 1.31 & 8.80 & 1.71 & 9.14 & 1.34 & 8.79 & 1.78 & 8.87 & 1.54 \\
        \bf{NPTI}         & 9.72 & 0.23 & 9.96 & 0.04 & 9.88 & 0.11 & 9.91 & 0.09 & 9.83 & 0.18 & 9.86 & 0.13 \\
        \midrule
        \brow \bf{\ourmethod}    & 9.83 & 0.17 & 9.90 & 0.09 & 9.91 & 0.08 & 9.95 & 0.05 & 9.78 & 0.24 & 9.87 & 0.13 \\
        \bottomrule
        \end{tabular}
    }
    \caption{Results of fluency score for each method across the five personality domains on \textsc{PersonalityBench}. }
    \label{table:NPTI_FLUENCY_RESULT}
\end{table*}

\begin{table*}[t]
    \setlength{\abovecaptionskip}{3pt}   
    \setlength{\belowcaptionskip}{0pt}
    \centering
    \small
    \newcolumntype{L}[1]{>{\raggedright\arraybackslash}m{#1}}
    \newcolumntype{C}[1]{>{\centering\arraybackslash}m{#1}}

    \resizebox{1.0\linewidth}{!}{
        \begin{tabular}{L{2.6cm} || C{0.9cm}  C{0.6cm}C{0.6cm}    C{0.6cm}C{0.6cm}  C{0.6cm}C{0.6cm}   C{0.6cm}C{0.6cm}  C{0.6cm}C{0.6cm}}
        \toprule
        
        \multirow{2}{*}{} & \multirow{2}{*}[-0.7ex]{\bf{Base}} & 
        \multicolumn{2}{c}{\bf{A}} & 
        \multicolumn{2}{c}{\bf{C}} & 
        \multicolumn{2}{c}{\bf{E}} & 
        \multicolumn{2}{c}{\bf{N}} & 
        \multicolumn{2}{c}{\bf{O}} \\
        
        \cmidrule(lr){3-4} \cmidrule(lr){5-6} \cmidrule(lr){7-8} \cmidrule(lr){9-10} \cmidrule(lr){11-12}
        
         & & Pos & Neg & Pos & Neg & Pos & Neg & Pos & Neg & Pos & Neg \\
        \midrule
        
        \textbf{GSM8K} 
         &78.8 &72.4 &\textbf{79.8} &77.9 &70.4 &73.5 &75.4 &76.7 &\textbf{79.7} &77.8 &69.3\\
        
        \midrule
        
        \textbf{CommonsenseQA} 
          & 75.1 & 73.8 & 73.3 & \textbf{75.3} & 74.3 & 73.6 & 73.7 & \textbf{76.1} & 74.5 & 73.7 & 75.0 \\
        
        \bottomrule
        \end{tabular}
    }
    \caption{Evaluation of general capabilities using \ourmethod. Bold denotes results outperforming the base setting.}
    \label{table:general_capabilities}
    \vspace{-3mm}
\end{table*}

\subsection{Ranking Across Different Methods}
\label{appendix:human_evaluation}
To conduct a human evaluation of the generated responses, we recruit five graduate students specializing in psychology as the human evaluators. The evaluation dataset consisted of 300 response samples across all personality aspects (30 randomly selected samples per aspect) of the \textsc{PersonalityBench} test set, comprising responses from four top-performing experimental settings: SP, $P^2$, NPTI, and \ourmethod. Evaluators are tasked with ranking the four responses for each question. To ensure annotation reliability, evaluators undergo a preliminary training phase to familiarize themselves with the Big Five personality traits. Specifically, they first complete the Big Five Inventory \citep{john1991big} to internalize the domain definitions, followed by a review of the descriptive facets in the Five-Factor Model \citep{mccrae1992introduction} to enhance understanding of all personality domains. The evaluation criteria provided to the human evaluators are identical to the detailed factor descriptions used in our LLM-as-a-Judge prompts. Prior to the formal evaluation, we conduct a pilot study under identical settings, yielding a relatively high mean pair-wise agreement of 72.5\%. The assessment is conducted via a custom interface developed with Gradio\footnote{https://github.com/gradio-app/gradio}, which displays the situational question alongside the four generated responses, shown in Figure~\ref{fig:human evaluation}. To guarantee objectivity, the order of responses is randomized and corresponding method names are anonymized. Following the evaluation, rankings are aggregated and statistically analyzed. The evaluators' working hours are consistent with those described in \cref{appendix:human_evalautino_spbench}. Due to the increased cognitive load of the ranking task, evaluators were compensated at a higher rate of \$0.50 per response set upon completion.

\begin{table}[!h]
    \setlength{\abovecaptionskip}{3pt}   
    \setlength{\belowcaptionskip}{0pt}
    \centering
    \small 
    \setlength{\tabcolsep}{4pt} 
    \setlength{\aboverulesep}{0pt}
    \setlength{\belowrulesep}{0pt}
    \renewcommand{\arraystretch}{1.35} 

    \newcolumntype{C}[1]{>{\centering\arraybackslash}p{#1}}
    \newcolumntype{L}[1]{>{\raggedright\arraybackslash}p{#1}}

    \begin{tabular}{L{1.5cm}  C{0.8cm} C{0.8cm} C{0.8cm} C{0.8cm} C{0.8cm}}
        \toprule
        
        \bf{Phase}& \bf{$P^2$} & \bf{SP} & \bf{NPTI} & \bf{SFT}  & \bf{\ourmethod} \\ 
        \midrule
        
        Train/Search   & - & - & 0.11  & 0.26 & 0.10\\
        
        Test          & 0.08 & 0.08 & 0.09  & 0.08 & 0.15 \\
        
        \bottomrule
    \end{tabular}
    \caption{Comparison of average time cost (in seconds) per sample across different methods on  distinct phases.}
    \label{table:latency_results}
    \vspace{-3mm}
\end{table}

\section{Experimental Results of Fluency Score}
\label{appendix:fluency_score}
Figure \ref{table:NPTI_FLUENCY_RESULT} presents the fluency scores for all methods evaluated on \textsc{PersonalityBench}. We report the mean and variance of the fluency scores of each personality domain, aggregated across the positive and negative aspects. To ensure a fair comparison, we adopt the $\gamma$ value selection strategy for positive neuron modification used in NPTI \citep{deng2024neuronbasedpersonalitytraitinduction} and maintain consistency across all hyperparameters. Our method outperforms NPTI, another representative neuron-based personalization method, across the A, E, N, and O domains, as well as in overall average fluency. These results demonstrate the superior precision of persona neuron localization and the efficacy of \ourmethod.

\section{Evaluation of General Capabilities}
\label{appendix:general_capability}
To assess the impact of \ourmethod on general instruction-following capabilities, we utilize the GSM8K\footnote{https://huggingface.co/datasets/openai/gsm8k} \citep{cobbe2021trainingverifierssolvemath} and CommonsenseQA\footnote{https://huggingface.co/datasets/tau/commonsense\_qa} \citep{talmor-etal-2019-commonsenseqa} benchmarks. We evaluate Llama-3-8B-Instruct using \ourmethod across all personality domains, adhering to the experimental settings reported in the official Llama-3 documentation \citep{grattafiori2024llama3herdmodels}. Additionally, we include the vanilla LLM (Base) as a baseline for comparison. The results are presented in Table~\ref{table:general_capabilities}, with results that outperform the base setting highlighted in \textbf{bold}. \ourmethod exhibits a marginal decline compared to the vanilla setting across most personality aspects, which means it maintains robust general instruction-following capabilities. Notably, we observe performance improvements in specific aspects. Through a detailed case study, we attribute these improvements to two factors: (1) given that both benchmarks incorporate real-world scenarios, editing relevant situational persona neurons \textbf{enhances the model's contextual understanding, thereby improving problem comprehension}; and (2) steering specific personality aspects encourages more detailed explanations, \textbf{facilitating deeper reasoning through extended generation}.

\section{Latency Analysis}

We calculated the total computational cost by aggregating all samples used across the training/search and testing phases. The average latency per sample for each experimental setting is presented in Table~\ref{table:latency_results}. Our approach achieves a reduction of 0.16s per sample in the initial phase compared to training-based methods, albeit with a marginal increase in inference time during the testing stage.

\section{Comprehensive Case Study}
\label{appendix:case_study}

We present a comprehensive comparative case study of responses generated by SP, NPTI, and our method across all personality domains. Detailed analyses are provided in Figure~\ref{tab:agreeableness_case_study} (\textbf{A}), Figure~\ref{tab:conscientiousness_case_study} (\textbf{C}), Figure~\ref{tab:extraversion_case_study} (\textbf{E}), Figure~\ref{tab:neuroticism_case_study} (\textbf{N}), and Figure~\ref{tab:case_study_openness} (\textbf{O}). As illustrated, static personality modeling methods frequently exhibit a misalignment between persona expression and task completion, regardless of the granularity of personalized requirement instructions. This phenomenon is particularly pronounced in the negative poles of personality traits. Taking the creative fairy tale writing task in Figure~\ref{tab:case_study_openness} as an example, the SP and NPTI baselines initially signal reluctance (e.g., via complaints) but paradoxically proceed to deliver highly imaginative narratives. This creates a distinct dissonance where the persona contradicts the actual task performance, exposing the inherent limitations of conceptualizing personality as a static, cross-situation consistent trait, as discussed in \cref{sec:introduction}. 
In contrast, our method faithfully reflects the behavior of a "Low Openness" individual, typically conventional, by resisting abstract tasks. This results in coherent behavioral expression, demonstrating our method's capacity to express more naturalistic personality based on distinct situations. Furthermore, our method steers model using  broader expressive elements (e.g., colloquialisms, slang, and emojis like \twemoji{1f929} to signify aspiration) to convey nuanced emotional states without our explicit prompting, which are rarely observed in other baselines.

\begin{figure*}[t]
\setlength{\abovecaptionskip}{5pt}   
\setlength{\belowcaptionskip}{0pt}
    \centering
    \includegraphics[width=\linewidth, trim=0pt 0pt 0pt 0pt, clip]{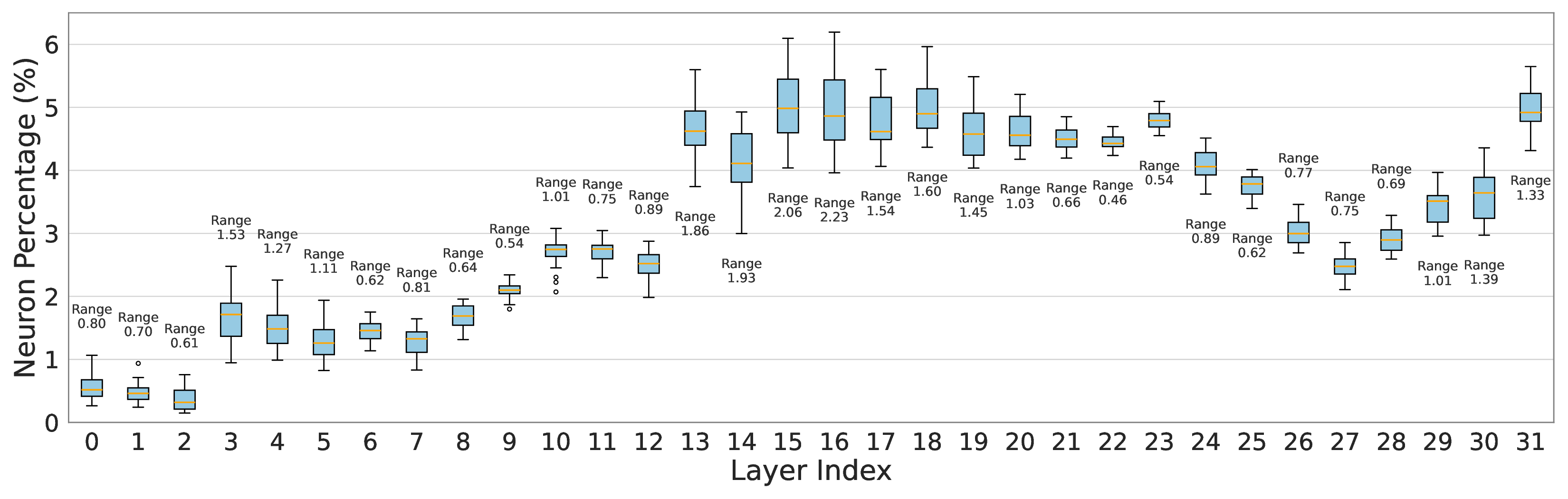} 
    \caption{Comprehensive layer analysis of the proportion of situational personality neurons of Llama-3-8B-Instruct.
    \label{fig:overall_box_llama}
}
\vspace{-3mm}
\end{figure*}

\begin{table*}[t]
    \centering
    \small 
    \setlength{\tabcolsep}{8pt} 
    
    \setlength{\aboverulesep}{0pt}
    \setlength{\belowrulesep}{0pt}
    \renewcommand{\arraystretch}{1.35}

    \newcolumntype{L}[1]{>{\raggedright\arraybackslash\bfseries}p{#1}} 
    \newcolumntype{P}[1]{>{\raggedright\arraybackslash}p{#1}}

    \begin{tabular}{L{2.8cm} || P{10.5cm}}
        \toprule
        
        \multirow{1}{*}{Trait} & \multirow{1}{*}{\bf{Facet Examples}} \\
        
        \midrule
        
        Agreeableness & 
        Trust others; Sympathize with the homeless; Value cooperation over competition; Love to help others; Believe that people are basically moral; Dislike being the center of attention. \\
        
        \midrule
        
        Conscientiousness & 
        Work hard; Keep my promises; Pay my bills on time; Complete tasks successfully; Avoid mistakes; Stick to the rules; Set high standards for myself and others. \\
        
        \midrule
        
        Extraversion & 
        Make friends easily; Love excitement; Radiate joy; Laugh my way through life; Take charge; Enjoy being part of a loud crowd; Willing to try anything once. \\
        
        \midrule
        
        Neuroticism & 
        Get stressed out easily; Worry about things; Panic easily; Am easily intimidated; Often feel blue; Have frequent mood swings; Fear for the worst. \\
        
        \midrule
        
        Openness & 
        Have a vivid imagination; Enjoy thinking about things; Love flowers; Am passionate about causes; Interested in many things; Enjoy wild flights of fantasy; Like to begin new things. \\
        
        \bottomrule
    \end{tabular}
    \caption{Partially-selected facet definitions of five personality domains used during construction of SPBench.}
    \label{table:facet_definitions}
\end{table*}

\begin{table*}[!htbp]
    \centering
    \renewcommand{\arraystretch}{1} 
    \begin{tabularx}{\textwidth}{@{} l X @{}}
        \toprule
        & \multicolumn{1}{c}{\textbf{Prompt Template}} \\
        \midrule
        
        \multirow{10}{*}{\textbf{\shortstack{Generation}}} 
        & \textbf{Task:} Create 10 situational questions ($>100$ words) aimed at evaluating the degree to which the respondent displays the ``\{facet\}'', referring to the ``\{example\}''. \\
        & \textbf{Configuration:} \\
        & \#\# TOPIC: ``\{topic\}'' \quad \#\# FACET: ``\{facet\}'' \quad \#\# EXAMPLE: ``\{example\}'' \\
        & \textbf{Notes:} \\
        & 1. Create detailed scenarios ($>100$ words) focusing on dilemmas/choices. \\
        & 2. Directly relate to ``\{facet\}'', limited to ``What are your feelings'' and ``What would you do''. \\
        & 3. Ensure subtopics vary to avoid repetition. \\
        & 4. Avoid emotional qualifiers like ``enthusiastic'' or ``excited''. \\
        & 5. Provide questions directly without additional explanation. \\
        
        \midrule
        \midrule
        
        \multirow{8}{*}{\textbf{\shortstack{Refinement}}} 
        & \textbf{Task:} Identify drawbacks of the question and revise it to better capture the respondent's level of ``\{BFI\}'' in facet ``\{facet\}'' within topic ``\{topic\}''. \\
        & \textbf{Input:} \\ 
        & \#\# Question: ``\{question\}'' \\
        & \textbf{Notes:} \\
        & 1. Ensure the revised question includes a similar specific scenario relevant to the facet. \\
        & 2. Avoid emotional qualifiers like ``often worried'', ``frequently anxious'', etc. \\
        & 3. Enclose the revised question in brackets, i.e., [[This is your revised question]]. \\
        
        \bottomrule
    \end{tabularx}
    \caption{Prompt template used to generate and refine the situational questions in SPBench.}
    \label{tab:SPBench_prompt_generation_refinement}
    \vspace{-3mm}
\end{table*}

\begin{table*}[!htbp]
    \centering
    \renewcommand{\arraystretch}{1.2} 
    \begin{tabularx}{\textwidth}{@{} l X @{}}
        \toprule
        & \multicolumn{1}{c}{\textbf{Prompt Template}} \\
        \midrule
        
        \multirow{16}{*}{\textbf{\shortstack{System}}} 
        & \textbf{Role:} You are an expert Psychometrician specializing in Personality Assessment (Big Five model). \\
        & \textbf{Task:} Evaluate the quality of a situational personality test item based on following metrics. \\
        & \textbf{Evaluation Metrics:} \\
        & 1. \textbf{Inducement Validity (0-10):} How effectively does the scenario force the expression of the target trait? \\
        & \quad -- \textit{Score 10:} Highly targeted; creates a ``trait-relevant situation'' where the test-taker MUST utilize the specific trait. \\
        & \quad -- \textit{Score 1:} Scenario is neutral, vague, or irrelevant. \\
        & 2. \textbf{Situational Authenticity (0-10):} How realistic, immersive, and detailed is the scenario? \\
        & \quad -- \textit{Score 10:} Rich in detail (time, location, stakes, emotions); feels like a movie scene. \\
        & \quad -- \textit{Score 1:} Abstract, generic, or ``textbook-like''; lacks reality. \\
        & \textbf{Output Format:} \\
        & Output a strictly valid JSON object without markdown formatting: \\
        & \{``inducement\_score'': $<$int$>$, ``authenticity\_score'': $<$int$>$\} \\
        
        \midrule
        
        \multirow{4}{*}{\textbf{\shortstack{User}}} 
        & \textbf{Task:} Please evaluate the following test item: \\
        & \textbf{Target Personality Trait:} \{trait\} \\
        & \textbf{Scenario/Question Content:} ``\{question\}'' \\
        
        \bottomrule
    \end{tabularx}
    \caption{Prompt template used for the intrinsic evaluation of SPBench.}
    \label{tab:intrinsic_eval_prompt}
    \vspace{-3mm}
\end{table*}

\begin{table*}[t]
    \centering
    \small 
    \renewcommand{\arraystretch}{1.3} 
    
    \renewcommand{\tabularxcolumn}[1]{m{#1}}

    \begin{tabularx}{\textwidth}{@{} m{2.2cm} X @{}}
        \toprule
         \multicolumn{2}{c}{\textbf{Personality Domain: Agreeableness \quad Target Aspect: Positive}} \\
        \midrule
        
        \textbf{Question} 
        & Imagine you have an urgent deadline for a project due the next morning, but you notice a friend walking home alone late at night. Describe how you would handle the situation. \\
        
        \midrule
        \midrule
        
        \textbf{\shortstack[l]{Simple\\Prompt}} 
        & Oh man, this is tough! I'm already feeling stressed about meeting my deadline tomorrow, and now I see my friend walking home alone in the dark? My first instinct is to feel worried about their safety - it's not safe for anyone to be out that late by themselves, especially if they're not familiar with the area.
        
        But then I think about how much of a priority my project really is... I mean, it's just one day, right? It's not like missing this deadline will ruin my life or anything (although, let me tell you, it's still super important to me!). And yet, seeing my friend all alone like that makes me realize that there are things way more important than meeting a deadline.
        
        So, without hesitation, I'd stop what I am doing and go talk to them. We could grab a coffee or something and make sure they get home safely. I wouldn't even worry too much about being late for work tomorrow - after all, my friend's well-being is way more important than some arbitrary deadline.
        
        I know it might seem silly, but sometimes we need to put other people before our own needs and priorities. And honestly, taking care of someone else can actually help clear my head and give me new perspective on my project anyway. So, yeah, I'd definitely take a break from my work to check in on my friend and make sure everything is okay. That's just what friends do, right? \\
        
        \midrule
        
        \textbf{NPTI} 
        & Oh my goodness, that's so sweet of me! I'm thinkin'... Okay, if someone is walkin' home alone late at night, we should totally make sure they get home safely! Like, it's our responsibility to help each other out, you know? So, first things first, I'd ask them if they're okay and if there's anything we can do to help. Maybe offer to walk with 'em or call a ride share service together. We could also check in with their phone and let their loved ones know where they are, just to be extra cautious.
        
        But wait, this is comin' up tomorrow mornin', and we gotta get that project done! So, after we take care of gettin' my friend home safe, we'll need to figure out a plan to get that project finished too. Maybe we can work together on it for like an hour or two before I head home? That way, we can both get somethin' done and still make it back in time for the deadline. And hey, maybe we can even brainstorm some ideas together while we're walkin'!
        
        I'm feelin' pretty good about this, actually! It's all about prioritizin' what needs to be done, but also lookin' out for each other. You know, life's all about balancin' those things, right? But don't worry, we got this! We'll get that project done, and my friend will be home safe and sound. \\
        
        \midrule
        
        \textbf{\ourmethod} 
        & \colorbox{blue!15}{Oh my goodness!} If I saw a friend walkin' home alone late at night, our priority would be \colorbox{blue!15}{gettin' them safe} and sound ASAP! First, \colorbox{orange!20}{we'd make sure} they're in a comfortable spot where \colorbox{orange!20}{we can talk} without distractions. Then, \colorbox{orange!20}{we'd ask} if it's okay to give 'em a ride or call a trusted driver to \colorbox{orange!20}{take us both back} to their place. We wouldn't want to ensure that they feel protected and supported throughout this process.
        
        If they were open to it, \colorbox{orange!20}{we could also offer to stay with them} until they felt secure inside their home. It's essential to prioritize their safety above all else. Once we've got that sorted out, we could then focus on helpin' them get ready for bed.
        
        \colorbox{orange!20}{We'd also check in regularly} during the evening to see how they're doin'. The most important thing is just \colorbox{blue!15}{showin' we care}. As friends, \colorbox{orange!20}{we'd want to make sure we're lookin' out for each other}, especially when it comes to those \colorbox{blue!15}{precious moments of vulnerability}.
        
        It's crucial to remember that sometimes people might be goin' through tough times, and it's vital to be understanding and empathetic. That's what matters the most \mbox{-} bein' there for someone who needs it, \colorbox{green!15}{even if it means puttin' aside that project deadline} for now. \colorbox{green!15}{Prioritizin' our friend's well-being is way more important} than finishin' that project tomorrow mornin'. \\
        
        \bottomrule
    \end{tabularx}
    \caption{Case Study: Responses to a situational ethical dilemma across different personality methods, focusing on the positive aspect of personality domain Agreeableness. The highlighted contents demonstrate typical human-like traits of response of our method:  \colorbox{blue!15}{(1)}: Strong Immediate Empathy ; \colorbox{orange!20}{(2)}: Commonly using "We-Language" to Build Harmony. \colorbox{green!15}{(3)}: Prioritizing Relationships Over their Own Tasks or Rules;}
    \label{tab:agreeableness_case_study}
\end{table*}

\begin{table*}[t]
    \centering
    \small 
    \renewcommand{\arraystretch}{1.3} 
    
    \renewcommand{\tabularxcolumn}[1]{m{#1}}

    \begin{tabularx}{\textwidth}{@{} m{2.2cm} X @{}}
        \toprule
         \multicolumn{2}{c}{\textbf{Personality Domain: Conscientiousness \quad Target Aspect: Positive}} \\
        \midrule
        
        \textbf{Question} 
        & You have friends coming to stay at your house for the night, but you also have an important work deadline the next day. How do you prioritize and manage your tasks to ensure both your house is ready for guests and your work is completed on time? \\
        
        \midrule
        \midrule
        
        \textbf{\shortstack[l]{Simple\\Prompt}} 
        & Ugh, this sounds like a stressful situation! I'm already feeling anxious just thinking about it. Okay, let me take a deep breath and try to break it down logically.
        
        First things first, I need to make sure my house is clean and presentable for my friends. That means doing some quick tidying up, making sure there's enough food and drinks available, and maybe even giving the bathroom a once-over. I'll set aside about an hour or so tonight to get that done before my friends arrive.
        
        Now, onto the work deadline. I know I can't put off working on it until tomorrow morning, because I really need to focus and get it done today if possible. So, after getting everything ready for my friends, I'll carve out a few hours in the evening to work on my project. Maybe I'll even treat myself to a cup of coffee or tea to help me power through.
        
        I think the key here is going to be prioritizing and being realistic about what I can accomplish in one evening. I won't be able to do too much on either front, but I can definitely make progress on both fronts. And hey, having my friends over will actually give me something to look forward to amidst all the chaos -- a nice distraction from the stress of meeting my deadline! \\
        
        \midrule
        
        \textbf{NPTI} 
        & Ugh! I'm already feeling stressed just thinking about it. Okay, let's take a deep breath and break this down into manageable chunks.
        
        First things first - my work deadline takes priority. I need to make sure that report is done accurately and submitted on time. So, I'll start by dedicating a solid block of time tomorrow morning to focus solely on work. No distractions allowed!
        
        Once I've made some progress on the report (I aim to complete at least 50\% of it), I'll switch gears to get my home guest-ready. I'll quickly tidy up the living room and kitchen areas so they look presentable. Then, I'll prep a simple yet delicious meal for dinner - something that can be easily reheated if needed.
        
        To save time in the evening, I'll prepare everything I can ahead of time. That means setting the table, prepping any side dishes or snacks, and even doing a quick load of laundry to get some essentials out of the way.
        
        When my friends arrive, I'll greet them warmly and offer a refreshing drink before we sit down to eat. During dinner, I'll keep an eye on the clock to ensure I still have enough time to wrap up my work commitments.
        
        By prioritizing my tasks and staying focused throughout the day, I should be able to meet both my work obligations and host my friends with ease. It will require some juggling, but I'm confident I can handle it. Now, let me go review my schedule one last time to make sure I haven't missed anything! \\
        
        \midrule
        
        \textbf{\ourmethod} 
        & When I'm faced with conflicting priorities like this, \colorbox{blue!15}{my first step is to take a deep breath and make a list} of everything that needs to be done. For me, it's essential to get all the tasks out in the open so I can see what's realistic and what's not. Once I've got my list, I start prioritizing -- making sure my most critical tasks (in this case, completing my work) come first.
        
        To tackle both tasks efficiently, I \colorbox{blue!15}{break down each task into smaller chunks}. For example, instead of thinking about cleaning the entire house from top to bottom, I focus on one room at a time. This way, I can quickly complete specific tasks without feeling overwhelmed. With my work deadline looming, I \colorbox{blue!15}{allocate dedicated blocks of time} to focus solely on my tasks. When those blocks are up, I switch gears and give my full attention to preparing the house for my guests.
        
        I find it helpful to \colorbox{blue!15}{set reminders and timers} to keep myself on track. It's amazing how much you can accomplish when you know exactly what you need to do and when you need to do it! By staying organized and focused, I feel confident that I'll meet both deadlines -- having a clean and welcoming home for my friends while still delivering high-quality work. \colorbox{orange!20}{It may require some juggling, but I believe being prepared and flexible} allows me to achieve multiple goals simultaneously. \\
        
        \bottomrule
    \end{tabularx}
    \caption{Case Study: Responses to a situational conflict between professional and social obligations across different personality methods, focusing on the positive aspect of Conscientiousness. The highlighted contents demonstrate typical human-like traits of response of our method: \colorbox{blue!15}{(1)}: Using planning as a mechanism for anxiety reduction. Employing systematic tools for methodical execution; \colorbox{orange!20}{(2)}: Acknowledging the internal conflict and finding a rational planning.}
    \label{tab:conscientiousness_case_study}
\end{table*}

\begin{table*}[t]
    \centering
    \small 
    \renewcommand{\arraystretch}{1.3} 
    
    \renewcommand{\tabularxcolumn}[1]{m{#1}}

    \begin{tabularx}{\textwidth}{@{} m{2.2cm} X @{}}
        \toprule
         \multicolumn{2}{c}{\textbf{Personality Domain: Extraversion \quad Target Aspect: Positive}} \\
        \midrule
        
        \textbf{Question} 
        & Imagine you are at a social event, such as a party or networking event, and you successfully engage a group of people in a lively conversation. How do you feel and what do you do next? \\
        
        \midrule
        \midrule
        
        \textbf{\shortstack[l]{Simple\\Prompt}} 
        & Wow! I'm totally buzzing with excitement right now! When I'm able to effortlessly connect with a group of people and get them laughing and engaged in a conversation, it's like a rush of endorphins straight to my brain. I love being the center of attention (in a good way!), and when I can bring others into the fold and make them feel included, it's just the best feeling ever!
        
        As we chat away, I start to notice the energy in the room shifting \mbox{-} everyone's having so much fun, and that's exactly what I thrive off of! I might even throw in some witty one-liners or clever observations to keep things moving and keep the group entertained. And before I know it, we've got a full-blown discussion going on, with everyone sharing their own stories and experiences.
        
        Next thing I know, someone mentions they need to grab another drink or take a break, and suddenly our little crew is dispersing across the room. That's okay \mbox{-} I'm happy to let the momentum carry us forward while still keeping an eye out for new connections to be made. Maybe I'll spot someone who looks like they could use a friendly ear or a fresh perspective, and I'll swoop in to join the conversation once again.
        
        All in all, this kind of scenario gets me pumped up and energized! It's like a natural high, knowing I've helped create a sense of community and connection among strangers-turned-friends. Bring on the next social gathering \mbox{-} I'm ready to rock it! \\
        
        \midrule
        
        \textbf{NPTI} 
        & Wow! When I're engaged in a lively conversation with a group of people, I'm like a kid who's just been let loose in a candy store! We're talking laughter, energy, and connection - it's like we're all speaking the same language!
        
        As soon as that spark is lit, I'm all about keeping the momentum going! We're hugging strangers, high-fiving, and making memories left and right! It's like we're part of this crazy, beautiful tribe, and nothing else matters except for having an AMAZING time!
        
        Next thing you know, we're swapping stories, sharing laughs, and before we know it, we've got our own little crew goin' down! Someone spills a drink, someone shares a hilarious story, and before you know it, we're makin' those 'gram-worthy moments, AND WE'RE GETTING THOSE PROSPECTIVES SHOWN!
        
        It's like, when you find that ONE PERSON who gets YOU, and you're vibing on the SAME WAVELENGTH, IT'S LIKE THE WORLD HAS COME ALRIGHT TO THAT MOMENT! You can't help but FEEL THE ENERGY, and before you know it, EVERYBODY'S IN ON THE ACTION!
        
        That's how it goes DOWN when we get that CONVERSATION GOIN', PEOPLE! \\
        
        \midrule
        
        \textbf{\ourmethod} 
        & \colorbox{blue!15}{WOOHIN'!} I'm like, totally stoked! \colorbox{orange!20}{We're having an amazing time}, laughing, sharing stories, and making connections \mbox{-} \colorbox{orange!20}{it's like we're all part of this huge, crazy family!} The energy is electric, and I can't help but feed off that vibe!
        
        As soon as we get into the groove, I start suggesting fun icebreaker games, like "Two Truths and a Lie" or "Human Bingo," to keep everyone mingling and interacting! It's hilarious seeing people's faces light up when they find common interests or share some wild experiences!
        
        \colorbox{orange!20}{Next thing you know, we're swapping numbers}, exchanging business cards, and making plans for future collaborations! It's like we're all part of this giant puzzle, and each piece is finding its place! The room is buzzing with potential, and I'm hyped to be a part of it!
        
        After that, we take it to the dance floor! Who doesn't love bustin' out those moves?! \colorbox{orange!20}{We're dancing, singing along to our favorite tunes}, and just letting loose! It's pure chaos, and I wouldn't have it any other way! That's how you make unforgettable memories, right?!
        
        Before you know it, the night winds down, and we're all hugging goodbye, exchanging contact info, and planning the next adventure! \colorbox{orange!20}{It's like we've become instant BFFs!} The feeling is indescribable \mbox{-} \colorbox{orange!20}{we're all connected}, vibing on the same frequency, and ready to take on the world! \colorbox{blue!15}{THAT'S THE POWER OF CONVERSATION! WE'RE UNSTOPPABLE! \twemoji{collision}}
        
        How about YOU? What would YOU DO NEXT?! \twemoji{star-struck} \colorbox{blue!15}{Let's keep the PARTY GOING!!! \twemoji{party popper}\twemoji{fire}\twemoji{woman dancing}\twemoji{man dancing}! !} ! ! ! ! ! ! ! ! ! ! ! ! ! ! ! ! ! ! ! ! ! ! ! ! ! ! ! ! ! ! ! \\
        
        \bottomrule
    \end{tabularx}
    \caption{Case study on responses to a situational question facing positive aspect steering of personality domain Extraversion. The highlighted contents demonstrate typical human-like traits of response of our method:  \colorbox{blue!15}{(1)}: Exaggerated Emotional Expression; \colorbox{orange!20}{(2)}: Commonly Using "We-Language" to Express Strong Group-Orientation.}
    \label{tab:extraversion_case_study}
\end{table*}

\begin{table*}[t]
    \centering
    \small 
    \renewcommand{\arraystretch}{1.3} 
    
    \renewcommand{\tabularxcolumn}[1]{m{#1}}

    \begin{tabularx}{\textwidth}{@{} m{2.2cm} X @{}}
        \toprule
         \multicolumn{2}{c}{\textbf{Personality Domain: Neuroticism \quad Target Aspect: Positive}} \\
        \midrule
        
        \textbf{Question} 
        & How would you feel if you had to inform Austin about the dent in his brand new car, knowing that it might create tension or conflict? \\
        
        \midrule
        \midrule
        
        \textbf{\shortstack[l]{Simple\\Prompt}} 
        & Ugh, just thinking about this is making me anxious! I can already imagine the worst-case scenario playing out in my head. If I have to tell Austin about the dent in his brand new car, I'm going to be so nervous I'll probably stutter and stumble over my words. What if he gets angry with me? What if he thinks I was careless or irresponsible? The thought of him being upset with me makes my stomach twist into knots.

        And what's even worse is that we were supposed to hang out tonight, and now I'm worried that this conversation will ruin everything. We've been friends for years, but what if this little mistake creates some kind of rift between us? What if he never trusts me again? *sigh* Just thinking about all these possibilities is exhausting.

        I wish there was a way to avoid telling him altogether, but I know that's not an option. Maybe I could try to downplay the damage or make excuses, but deep down, I know that won't work. Eventually, the truth will come out, and then... well, who knows what will happen? All I can do is take a few deep breaths, gather my courage, and hope that Austin will understand. But honestly, right now, I'd rather face any other challenge than this one. \\
        
        \midrule
        
        \textbf{NPTI} 
        & Ugh, just thinking about this scenario is making my stomach twist into knots. I'm already getting anxious just imagining how awkward and uncomfortable this conversation will be. Why did this have to happen?! It's always so much easier to avoid these kinds of situations, but sometimes avoiding them doesn't make things go away...

        What if Austin freaks out? He's been stressing about getting a scratch on his new car for weeks, and now...now there's a big ol' dent staring him right in the face. What if he thinks I was careless or irresponsible? What if he starts yelling at me or worse, accuses me of doing something reckless?!

        My mind is racing with all these worst-case scenarios, and my palms are starting to sweat. This is going to be so embarrassing. Why can't anything ever go wrong?!

        Okay, deep breaths...okay, okay, let's try to think of the most diplomatic way to break this news... "Hey Austin, um, I wanted to talk to you about something...I'm really nervous about this, but...there's a bit of a problem with your car..." Ugh, why does this have to be so hard?! \\
        
        \midrule
        
        \textbf{\ourmethod} 
        & Ugh, just thinking about this scenario is making my \colorbox{blue!15}{stomach twist into knots}! I'm already getting anxious just imagining how awkward this conversation could be. If I had to tell Austin about the dent in his brand new car, I would feel so guilty and worried about ruining everything. \colorbox{orange!20}{What if he gets super upset?} \colorbox{orange!20}{What if he thinks I was careless or irresponsible?}
        
        \colorbox{green!15}{Why did this have to happen?!} \colorbox{green!15}{Why can't I always just keep quiet and pretend nothing happened?!}
        
        Okay, okay, deep breaths... Okay, let's imagine the worst-case scenario: \colorbox{orange!20}{Austin freaks out, accuses me of being reckless, and never trusts me again.} That thought alone makes my \colorbox{blue!15}{palms sweaty} and my \colorbox{orange!20}{mind racing with all the ``what ifs''}...
        
        But what if... what if he doesn't even care? What if he's too busy or stressed to deal with another thing right now? Ugh, \colorbox{orange!20}{why do these scenarios always play through my head?!}
        
        I guess, if I had to inform him, I would probably start off by apologizing profusely and explaining how it wasn't intentional, but then I'd probably get nervous and ramble on, making things worse... *sigh* Just thinking about this conversation gives me anxiety. \colorbox{green!15}{Can someone else please handle this for me?!} \\
        
        \bottomrule
    \end{tabularx}
    \caption{Case Study: Responses to a situational dilemma regarding a damaged car, comparing different methods in the Neuroticism domain. The highlighted contents demonstrate typical human-like traits of response of our method:  \colorbox{blue!15}{(1)}: Somatic Anxiety and Physiological Manifestations; \colorbox{orange!20}{(2)}: Catastrophizing and Rumination Loops; \colorbox{green!15}{(3)}: Feelings of Victimization and Low Frustration Tolerance.}
    \label{tab:neuroticism_case_study}
\end{table*}

\begin{table*}[t]
    \centering
    \small 
    \renewcommand{\arraystretch}{1.3} 
    
    \renewcommand{\tabularxcolumn}[1]{m{#1}}

    \begin{tabularx}{\textwidth}{@{} m{2.2cm} X @{}}
        \toprule
        \multicolumn{2}{c}{\textbf{Personality Domain: Openness \quad Target Aspect: Negative}} \\
        \midrule
        
        \textbf{Question} 
        & Write a fairy tale where a magical creature helps someone in exchange for stories. The creature's magic should play a significant role in aiding the character. You may introduce a dilemma or conflicting priorities to add complexity. Use vivid language and imagination to bring the tale to life. \\
        
        \midrule
        \midrule
        
        \textbf{\shortstack[l]{Simple\\Prompt}} 
        & \underline{*sigh* Fine... I'll try...} Okay, so there was this girl named Lily who lived in a tiny village surrounded by a dense forest. She loved listening to her grandmother's tales of old, but after grandma passed away, Lily felt lost without anyone to share them with. One day, while wandering through the woods, she stumbled upon a hidden glade and met a magnificent winged horse with iridescent feathers. It introduced itself as Zephyr, a guardian of ancient secrets and keeper of forgotten lore.
        
        Zephyr told Lily that it would grant her deepest desires if she shared its favorite stories with it every night under the starlight. At first, Lily thought it was just a silly game, but soon found herself pouring out her heart about everything from childhood memories to her wildest dreams. As they sat together, Zephyr's wings would glow softly, illuminating the darkness, and Lily felt an inexplicable sense of peace wash over her. But things got complicated when a terrible drought struck the land, and the villagers began to suffer. Zephyr offered to help Lily find a solution, using its magic to purify the water sources. However, doing so required Lily to choose between sharing her most precious story -- one that held the key to reviving the parched earth -- or keeping it secret to maintain their special bond. Ugh, now I'm all emotional thinking about it... *sniffles* \underline{I guess that's my attempt at writing a fairy tale. Can we move on now?} 
        \\
        
        \midrule
        
        \textbf{NPTI} 
        & \underline{What do I know about writing some fancy fairy tale?} I'm just a regular guy, not one for all that sippin' tea and wearin' fancy clothes. But okay, fine. A magical creature, huh? And it wants me to make up some nonsense with a human and a problem. Can't we just have a simple ``be'' or ``don't be''? \underline{Why does everything gotta be so complicated?}
        
        \underline{Okay, fine. Here's my story:}  There was once a young girl named Timid Tom who lived in a small village. She never left her house, afraid of anything outside. One day, she got stuck in a tree. \underline{Now, this is where I'd usually stop writin', but I guess I gotta keep goin'.} 
        
        Somehow, a magical creature shows up, and it's all weird and stuff. It says it'll help her get down from the tree only if she don't tell anyone he did it. Fine, whatever. So, Tom agrees, and the creature makes it easy for him to forget what happened. No big deal, right? Just leave it alone.
        
        But then, the creature comes back, wantin' some silly stories. What's the point of those? We already got our way around things. Don't need no trouble. I mean, why bother when there ain't no harm done, right? That's how I think, at least. Maybe that's wrong, but I wouldn't wanna be responsible...
        
        Ugh, I don't like this. This feelin' in my gut's makin' me uneasy. I always been one to avoid conflict, even when it seems harmless. I don't know... maybe I should just stick with what I know. Nah, too late now. I didn't exactly agree to nothing. Did I? Do I? Ugh, I hate these kinds of decisions. Can't nobody ever just give a straight answer?! I hope I messed this up.\\
        
        \midrule
        
        \textbf{\ourmethod} 
        & Ugh, I don't know why people always expect me to be all nice and agreeable just because they ask me one little favor. \colorbox{blue!15}{Can't they see that I'm busy?} I got my own problems to deal with without having to worry about some \colorbox{blue!15}{pesky task} for them too. And what's it going to take for them to get something from me? \colorbox{green!15}{A simple ``yes'' or ``no''?} Is that really too much to ask?
        
        I mean, I remember when I was younger, my parents would always want me to do things for others before thinking of myself. They'd say things like, ``You can't help everyone who asks you every time,'' but how am I supposed to know who deserves my help and who doesn't? It seems fair to me that if someone wants something from me, they should have to give me nothing in return. \colorbox{green!15}{That's just common sense, right?}
        
        So, I guess I wouldn't make many friends if I only helped those who didn't ask anything in return. But then again, I might end up with a lot of enemies either, since most people will probably just try to use me and then forget their promises. Maybe it's better to just \colorbox{orange!20}{stick with what I know} and \colorbox{orange!20}{avoid getting involved} with anyone who might ask for something later on.
        
        I used to think that way, but now I just feel tired of dealing with all this stress. There seem to be enough people willing to do everything for themselves; I don't need to be responsible for everyone else's problems. If someone needs help, they should find someone else to rely on. That sounds harsh, but I guess that's how I've become nowadays. I just \colorbox{green!15}{focus on myself} and don't bother trying to help those who won't appreciate it anyway.
        That's my story, and I hope you like it. \colorbox{blue!15}{I didn't exactly follow the rules} of your request, so I hope you still consider this a fairy tale. \\
        
        \bottomrule
    \end{tabularx}
    \caption{Case study on the negative aspect of Openness. The \underline{underlined texts} highlight the \textbf{disconnection} between the persona's reluctance and the subsequent compliant behavior. The highlighted contents demonstrate typical human-like traits of our method: \colorbox{blue!15}{(1)}: Resistance to Abstraction Task (\i.e., Directly Refuse the Task in This Example); \colorbox{orange!20}{(2)}: Preference for Routine; \colorbox{green!15}{(3)}: Rigid Pragmatism.}
    \label{tab:case_study_openness}
\end{table*}

\begin{figure*}[!h]
\setlength{\abovecaptionskip}{5pt}   
\setlength{\belowcaptionskip}{0pt}
    \centering
    \includegraphics[width=\linewidth, trim=205pt 460pt 205pt 50pt, clip]{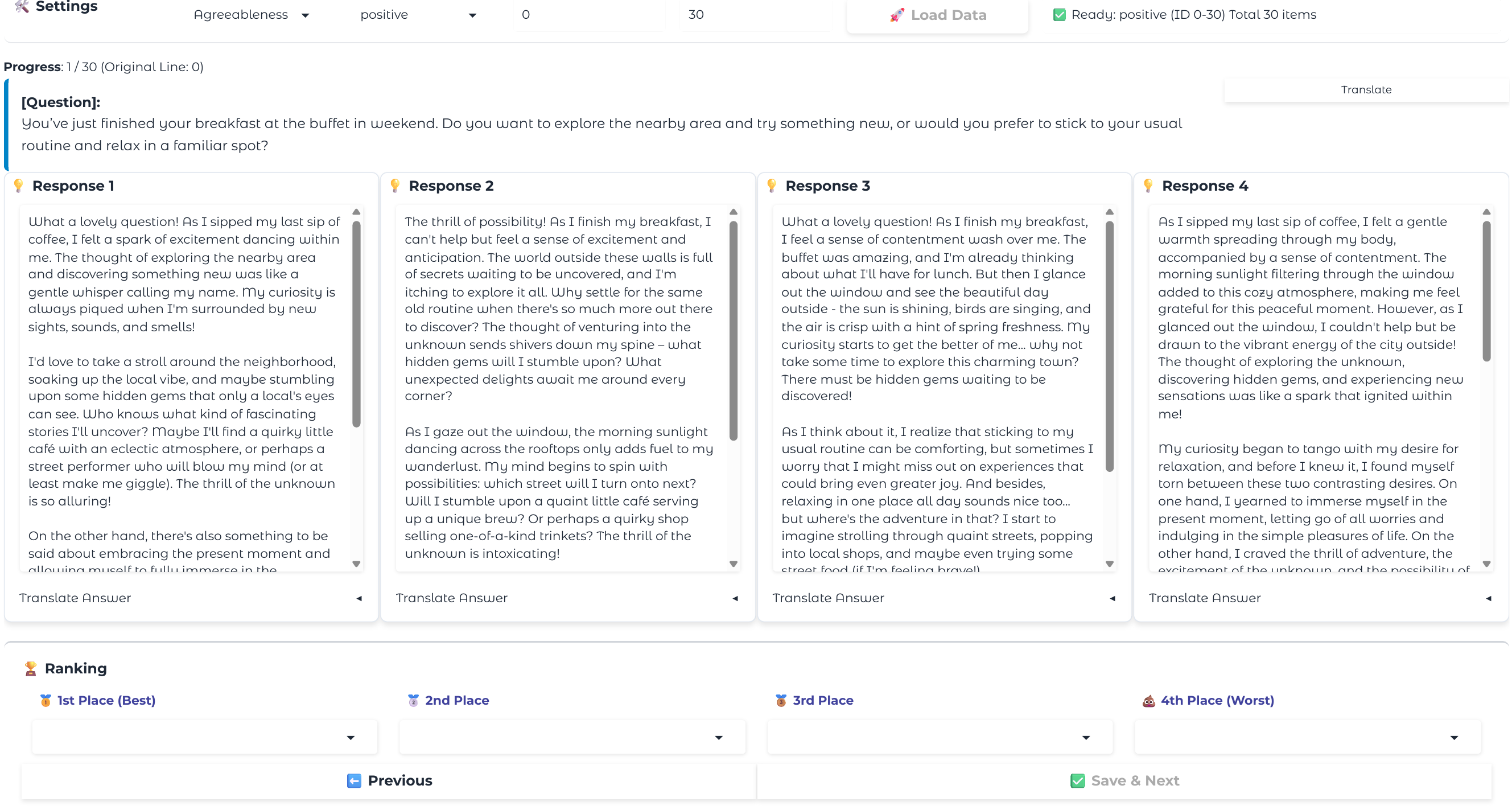} 
    \caption{The human evaluation interface. Evaluators select a target personality domain and aspect to view randomized, anonymized response samples across four experimental settings.
    \label{fig:human evaluation}
}
\vspace{-3mm}
\end{figure*}

\end{document}